\documentclass[sigconf]{acmart}
\usepackage{cleveref}
\usepackage{hyperref}
\usepackage{url}
\usepackage[utf8]{inputenc} 
\usepackage[T1]{fontenc}    
\usepackage{hyperref}       
\usepackage{url}            
\usepackage{booktabs}       
\usepackage{colortbl}
\usepackage{xcolor}
\usepackage{subcaption}
\usepackage{makecell}
\usepackage{amsfonts}       
\usepackage{nicefrac}       
\usepackage{microtype}      
\usepackage{xcolor}         
\usepackage{natbib}
\usepackage{graphicx}
\usepackage{wrapfig}
\usepackage{multirow}
\usepackage{epsfig}
\usepackage{amsmath}
\usepackage{marvosym}
\usepackage{color}
\usepackage{algorithm}
\usepackage{algpseudocode}
\usepackage{verbatim}
\usepackage{threeparttable} 
\usepackage{bm}
\usepackage{svg}

\usepackage{wrapfig}

\usepackage{xspace}
\usepackage{bm}
\usepackage{soul}
\usepackage{balance}
\usepackage[a-2b]{pdfx}
\usepackage{amsthm}

\newtheorem{problem}{Problem Statement}
\newtheorem{myThm}{Theorem}

\newcommand{\eg}{\emph{e.g.},\xspace}
\newcommand{\ie}{\emph{i.e.},\xspace}
\newcommand{\wrt}{\emph{w.r.t.}\xspace}

\newcommand{\sth}{\emph{s.t.},\xspace}

\newcommand\tabref[1]{Table~\ref{#1}}

\newcommand\appref[1]{Appendix~\ref{#1}}

\newcommand{\eat}[1]{}

\newcommand{\update}[1]{{\textcolor{black}{#1}}}
\newcommand{\boldres}[1]{{\textbf{\textcolor{red}{#1}}}}
\newcommand{\secondres}[1]{{\underline{\textcolor{blue}{#1}}}}

\newcommand{\model}{{\sc SDE-Mamba}\xspace}
\AtBeginDocument{%
  }

\copyrightyear{2025}
\acmYear{2025}
\setcopyright{acmlicensed}\acmConference[KDD '25]{Proceedings of the 31st ACM SIGKDD Conference on Knowledge Discovery and Data Mining V.2}{August 3--7, 2025}{Toronto, ON, Canada}
\acmBooktitle{Proceedings of the 31st ACM SIGKDD Conference on Knowledge Discovery and Data Mining V.2 (KDD '25), August 3--7, 2025, Toronto, ON, Canada}
\acmDOI{10.1145/3711896.3737119}
\acmISBN{979-8-4007-1454-2/2025/08}

\begin{document}

\title{SDE: A Simplified and Disentangled Dependency Encoding Framework for State Space Models in Time Series Forecasting}

\author{Zixuan Weng}
\orcid{0009-0000-6820-4286}
\affiliation{%
  \institution{The Hong Kong University of Science and Technology (Guangzhou)}
  \department{}
  \city{Guangzhou}
  \state{Guangdong}
  \country{China}
}
\email{zxweng0701@gmail.com}

\author{Jindong Han}
\orcid{0000-0002-1542-6149}
\affiliation{%
  \institution{The Hong Kong University of Science and Technology}
  \department{}
  \city{Hong Kong}
  \state{}
  \country{China}
}
\email{jhanao@connect.ust.hk}

\author{Wenzhao Jiang}
\orcid{0009-0006-1081-8684}
\affiliation{%
  \institution{The Hong Kong University of Science and Technology (Guangzhou)}
  \department{}
  \city{Guangzhou}
  \state{Guangdong}
  \country{China}
}
\email{wjiang431@connect.hkust-gz.edu.cn}

\author{Hao Liu}
\authornote{Hao Liu is the corresponding author.}
\orcid{0000-0003-4271-1567}

\affiliation{%
  \institution{The Hong Kong University of Science and Technology (Guangzhou)}
  \city{Guangzhou}
  \state{Guangdong}
  \country{China}
}

\affiliation{%
  \institution{The Hong Kong University of Science and Technology}
  \city{Hong Kong}
  \country{China}
}

\email{liuh@ust.hk}

\renewcommand{\shortauthors}{Zixuan Weng et al.}

\begin{abstract}
In recent years, advancements in deep learning have spurred the development of numerous models for Long-term Time Series Forecasting (LTSF). However, most existing approaches struggle to fully capture the complex and structured dependencies inherent in time series data. In this work, we identify and formally define three critical dependencies that are fundamental to forecasting accuracy: order dependency and semantic dependency along the temporal dimension, as well as cross-variate dependency across the feature dimension. These dependencies are often treated in isolation, and improper handling can introduce noise and degrade forecasting performance. To bridge this gap, we investigate the potential of State Space Models (SSMs) for LTSF and emphasize their inherent advantages in capturing these essential dependencies. Additionally, we empirically observe that excessive nonlinearity in conventional SSMs introduce redundancy when applied to semantically sparse time series data. Motivated by this insight, we propose SDE (Simplified and Disentangled Dependency Encoding), a novel framework designed to enhance the capability of SSMs for LTSF. Specifically, we first eliminate unnecessary nonlinearities in vanilla SSMs, thereby improving the suitability for time series forecasting. Building on this foundation, we introduce a disentangled encoding strategy, which empowers SSMs to efficiently model cross-variate dependencies while mitigating interference between the temporal and feature dimensions. Furthermore, we provide rigorous theoretical justifications to substantiate our design choices. Extensive experiments on nine real-world benchmark datasets demonstrate that SDE-enhanced SSMs consistently outperform state-of-the-art time series forecasting models. 
\end{abstract}

\ccsdesc[500]{Information systems~Spatial-temporal systems}
\ccsdesc[500]{Computing methodologies~Artificial intelligence}



\keywords{Time series forecasting; Disentangled encoding; State space model}


\maketitle

\newcommand\kddavailabilityurl{https://doi.org/10.5281/zenodo.15564796
}

\ifdefempty{\kddavailabilityurl}{}{
\begingroup\small\noindent\raggedright\textbf{KDD Availability Link:}\\
The source code of this paper has been made publicly available at \url{https://github.com/YukinoAsuna/SAMBA}.
\endgroup
}

\section{Introduction}

Long-term Time Series Forecasting (LTSF) plays a crucial role in numerous real-world applications, such as early disaster warning systems and long-term energy planning~\cite{zhou2021informer,wu2021autoformer,yeh2024rpmixer,wang2024language}. The effectiveness of LTSF models relies heavily on their ability to capture meaningful dependencies from extensive historical observations. As a result, traditional approaches often leverage periodicity and trend patterns to guide model design and enhance interpretability~\cite{wu2021autoformer}. However, the recent success of linear models~\cite{DLinear} and patching strategies in Transformers~\cite{patchtst} suggests that existing formulations may overlook deeper and more essential factors that drive predictability in time series.

In this work, we revisit the foundations of time series modeling by decomposing temporal dependencies into three key aspects: (i) order dependency, which reflects the sequential structure across time steps; (ii) semantic dependency, which captures high-level temporal patterns beyond surface numerical values; and (iii) cross-variate dependency, which models interactions among variables in multivariate time series. We argue that effective LTSF requires careful modeling of all three dependencies. This provides a unified lens for analyzing existing models and their limitations.

Currently, deep LTSF models typically fall into three paradigms: (i) linear- and MLP-based models \cite{RLinear,DLinear,tide,chen2023tsmixer}, (ii) Transformer-based models \cite{wu2021autoformer,zhou2021informer,liu2023itransformer,patchtst,liu2024generative}, and (iii) State Space Models (SSMs) (e.g. RNN-based models)~\cite{lin2023segrnn,jia2024pgn,jhin2024addressing}. The majority of existing LTSF approaches rely on either MLP/linear models or Transformer-based architectures. However, both of them exhibit fundamental limitations in capturing order and semantic dependencies. Linear models (e.g., DLinear~\cite{DLinear}) excel at preserving temporal order through point-wise mappings, but struggle to capture semantic-level patterns due to the restricted capacity. Transformer models~\cite{transformer}, in contrast, are effective in learning semantic dependency via attention mechanisms, especially when enhanced with patching strategies~\cite{patchtst}, but often lack the structural inductive bias needed to preserve sequential order~\cite{DLinear,tan2024language}. Despite the use of positional encodings, the self-attention mechanism remains permutation-invariant, limiting its temporal sensitivity~\cite{tan2024language}.


On the other hand, SSMs offer a promising alternative. Their recursive structure naturally preserves order dependencies~\cite{gu2021combining}, and recent advances in deep SSMs have significantly improved their capacity to capture semantic patterns. Nevertheless, conventional SSM-based implementations, particularly through Recurrent Neural Networks (RNNs), still suffer from gradient vanishing and limited parallelism when handling long-range sequential dependencies, making them less competitive for LTSF tasks~\cite{jia2024pgn,lin2023segrnn}.


To address these issues, two main research directions emerged: (i) improving the efficiency and stability of RNN-style architectures~\cite{lin2023segrnn,jia2024pgn}, and (ii) developing new SSM variants tailored for long-range dependencies, such as the recently proposed Mamba~\cite{gu2023mamba,mambaNLP}. While these methods show promise in various sequence tasks, the direct application to LTSF faces two non-trivial challenges: (1) \textbf{Order-semantic dependency rebalance}: Mamba and similar SSMs are primarily designed for language modeling, where semantic richness dominates. In contrast, time series are often low in information density~\cite{DLinear} and require stronger modeling of order dependency. Naively applying such architectures may lead to imbalance, overfitting to semantic noise while ignoring essential sequential patterns. (2) \textbf{Cross-variate dependency modeling}: Multivariate forecasting relies on capturing variable interactions. Most SSMs lack explicit mechanisms for modeling cross-variate dependencies~\cite{traffic, liu2020polestar}. However, existing cross-variate modeling strategies (e.g., Channel-Dependent (CD) methods) often underperform compared to Channel-Independent (CI) approaches~\cite{liu2023itransformer}. This discrepancy arises from the over-smoothing effect in CD models, losing the unique temporal features of individual variables~\cite{CCM}. While hybrid designs (e.g., CI-then-CD) have been proposed~\cite{liu2023itransformer}, our empirical and theoretical analysis demonstrate that they still introduce redundant information from weakly correlated variables, ultimately compromising forecasting performance.

To tackle the above challenges, we propose \textbf{SDE}, a \textbf{S}implified and \textbf{D}isentangled Dependency \textbf{E}ncoding framework to enhance SSMs for LTSF. Specifically, SDE consists of two major components: simplification and disentangled encoding strategy. (1) Simplification mechanism: We eliminate unnecessary nonlinear components from existing SSMs, such as RNNs and Mamba. Our empirical and theoretical analysis show that such nonlinearities often cause overfitting in time series with low semantic complexity. By simplifying these components, we improve generalization and robustness across datasets. (2) Disentangled encoding strategy: We introduce a \textbf{theoretically grounded} disentangled encoding strategy that explicitly separates temporal and cross-variate dependencies. This disentangled design reduces mutual interference between dimensions and is universally applicable across both SSM and Transformer backbones, enabling flexible integration. Through extensive experiments on multiple real-world datasets, we demonstrate that our method—SDE-Mamba, a Mamba variant enhanced by our framework—consistently achieves state-of-the-art performance. Notably, SDE-Mamba adapts well to datasets with varying levels of cross-variate correlation, underscoring its robustness and generality. Our major contributions are summarized as follows:

\begin{itemize}
    \item 
    We systematically identify and formally define three critical types of dependencies in time series data, \ie order dependency, semantic dependency, and cross-variate dependency, to guide the design of LTSF models.
    \item We derive two key insights: (i) SSMs, compared to linear models and Transformers, exhibit superior capability in capturing both order and semantic dependencies; (ii) vanilla SSMs suffer from overfitting when applied to LTSF due to excessive nonlinear operations.
    \item In response, we propose the SDE framework, which consists of two core components: (i) a simplification mechanism applicable to both RNN- and Mamba-based SSMs, removing excessive nonlinear operations to improve generalization; and (ii) a \textbf{theoretically grounded} disentangled encoding strategy that effectively integrates cross-variate dependencies while preserving the dynamics of individual variates.
    \item Extensive experiments validate the effectiveness of our approach, with SDE-Mamba achieving SOTA performance across multiple LTSF benchmarks. Furthermore, our analysis showcase the broad applicability of disentangled encoding strategy, offering a generalizable framework for future LTSF model development.
\end{itemize}

\section{Related Works}

Deep learning has made significant advances in the field of natural language processing \cite{bert}, intelligent Internet of Things \cite{xiao2023know}, speech recognition \cite{speech_trans}, inspiring researchers to repurpose these models for time series forecasting \cite{zhou2021informer}; \cite{wu2021autoformer}. The mainstream methods for LTSF currently include linear models and Transformer-based models. In particular, because of the strong modeling capability of the transformer, numerous works have utilized transformers to achieve superior performance. However, applying Transformers to long sequence forecasting presents two major challenges: 1) computational overhead due to quadratic space-time complexity and 2) the difficulty of capturing long-term dependencies. Early approaches concentrated on modifying the architectural design to reduce complexity. For example, LogTrans \cite{logtrans} introduced convolutional self-attention to capture local information while utilizing LogSparse techniques to reduce complexity. Informer \cite{zhou2021informer} leveraged a self-attention distillation mechanism to extend the input length and proposed ProbSparse self-attention to optimize computation efficiency. Furthermore, Autoformer \cite{wu2021autoformer} selected a decomposed structural framework to disintegrate time and replaced traditional self-attention mechanisms with autocorrelation mechanisms to reduce complexity.

Recently, linear models have demonstrated superior performance with fewer parameters and higher efficiency, prompting researchers to question whether transformers are suitable for long sequence forecasting \cite{DLinear}.  Researchers have begun exploring how to let transformers exhibit their powerful performance in scenarios like natural language processing and speech based on the inherent properties of time series.  For example, considering the distribution shift problem inherent in time series, methods like Non-stationary transformers \cite{Non_stationary} have applied techniques to stabilize and reduce the inherent distribution shift of time series.  Given the insufficiency of semantic information at individual time points, the PatchTST \cite{patchtst} model, by patching to extract local information, addresses the issue of insufficient short-time series semantic content, thus enriching the semantic content of each token.  


Recent research shows that utilizing the CI strategy to achieve promising results \cite{patchtst}, inspiring researchers to explore a new encoding method that can consider both intra-variate and inter-variate interactions \cite{chen2023tsmixer}. Crossformer \cite{crossformer} adopted a cross-variate encoding method and a two-stage attention mechanism that considers both time and variate relationships. iTransformer \cite{liu2023itransformer} applied a data inversion method, considering the time series corresponding to each variate as a token and then capturing the relationships between variates using self-attention mechanisms. CARD \cite{wang2024card} explicitly models both cross-time dependency and cross-variate dependency, where the latter directly mixes these dependencies at each model layer. As a result, CARD still suffers from the issue mentioned in the introduction, which is the inappropriate mixing of cross-time and cross-variate dependencies. 

Unlike previous studies, we conduct a comprehensive analysis of the advantages of state space models (SSMs) in LTSF and propose SDE, an enhanced framework that fully leverages their strengths. In contrast to existing approaches, SDE-Mamba explicitly states-of-the-art (SOTA) performance across datasets with varying degrees of intervariate dependencies.

\section{Problem Formulation}
\label{sec:problem}

\begin{problem}
\textbf{Long-term Time Series Forecasting (LTSF).}
The goal is to learn a mapping function $\mathcal{F}(\cdot)$ that forecasts the temporal evolution of $N$ variates in the future $S$ time steps based on the observations in the past $T$ time steps, \ie
\begin{align}
    \mathcal{F}: \mathbb{R}^{N \times T} \to \mathbb{R}^{N \times S}, \quad \mathbf{X} \mapsto \mathbf{Y},
\end{align}
\begin{align}
    \mathbf{X} = (\mathbf{x}_1, \dots, \mathbf{x}_T), \quad \mathbf{Y} = (\mathbf{x}_{T+1}, \dots, \mathbf{x}_{T+S}).
\end{align}
where $\mathbf{x}_{t}=(c_{t}^{1}, \ldots, c_{t}^{N}) \in \mathbb{R}^N$ denotes the states of $N$ variates at time step $t.$
\end{problem}

Accurate LTSF depends on effectively capturing the following three types of dependencies:

(1) \emph{Order dependency:} It refers to the temporal ordering relationships among sequentially observed data points of a variate. Formally, given observations $c_{t}$ and $c_{s}$ where $t < s$, we omit the superscript $i$ of $c_t$ for brevity. The order dependency is significant for prediction if 
\begin{align}
    I(c_t; c_s \mid t,s) > I(c_t; c_s),
\end{align}
where $I(\cdot; \cdot)$ ($I(\cdot; \cdot \mid \cdot)$) denotes the (conditional) mutual information \cite{infor_theory1999}.
For example, the increasing trend of daily temperatures in the past week helps predict that the trend will likely continue.

(2) \emph{Semantic dependency:} It refers to the latent semantic relationships between historical and future data points of a variate. It goes beyond the superficial temporal ordering information, which is more stable across temporal contexts, and requires more expressive models to extract. Formally, the semantic dependency is significant for prediction if there exists a permutation-invariant (nonlinear) function $\mathcal{S}(\cdot)$ that maps inputs to semantic space, \sth
\begin{align}
H(\bm{c}_{T+1:T+S} | \mathcal{O}(\bm{c}_{1:T}), \mathcal{S}(\bm{c}_{1:T}) ) < H(\bm{c}_{T+1:T+S} | \mathcal{O}(\bm{c}_{1:T})). 
\end{align}where $H(\cdot|X)$ denotes the conditional entropy of a random variate given variate $X$ and $\mathcal{O}(\cdot)$ includes additional order dependencies \wrt numerical inputs. For instance, periodic patterns facilitate more precise forecasting of seasonal temperature patterns on top of the potential temporal trends.

(3) \emph{Cross-variate dependency:} It refers to the complex relationships between variate $i$ and $j$. The cross-variate dependency is significant for predicting variate $i$ if  
\begin{align*}
    I(\bm{c}_{1:T}^i; \bm{c}_{T+1:T+S}^i \mid \bm{c}_{1:T}^j) 
    &> I(\bm{c}_{1:T}^i; \bm{c}_{T+1:T+S}^i).
\end{align*}
For instance, exploiting co-evolving temperature patterns between the target region and adjacent regions could improve the prediction in the target region.


\section{Empirical Exploration of State Space Model for LTSF}
\label{sec:empirical_explorations}
An effective LTSF model should be able to take advantage of the three types of dependencies defined in Section \ref{sec:problem}. We first conduct in-depth ablation experiments to evaluate four models: two classical LTSF models, Linear and Transformer, as well as two state space models in capturing these dependencies.

\subsection{Why State Space model? A Deep Dive into Its Suitability for LTSF}
\label{subsection:mamba}
To verify the applicability of SSMs to LTSF, we selected SegRNN \cite{lin2023segrnn} and Mamba \cite{gu2023mamba} as two representative SSMs to evaluate SSM's ability to capture sequential dependencies and semantic dependencies. To assess the models' ability to capture order dependency, we first make the following assumption.

\textbf{Assumption 1.} The performance of an effective order dependency learner is significantly influenced by the order of the input sequence \cite{DLinear}.

According to Assumption 1, a greater performance variation indicates a greater reliance on order dependency. In addition, we also consider the SOTA model, iTransformer as one of our baslines. It is worth noting that iTransformer encodes temporal relationships using a Linear Model and then uses a Transformer encoder to capture cross-variate dependencies. Ideally, it is expected to be sensitive to order as well. Our results in Table \ref{tab:order_dependency_1} show that Linear model, Mamba and SegRNN surpass the Transformer in terms of capturing order dependency. This is because the Linear model, as well as the state-space models Mamba and SegRNN, which process sequences recursively, are all permutation-variant, i.e., sensitive to the order. Their sensitivity to sequence order is stronger than that of the Transformer, which relies on the permutation-invariant self-attention mechanism. This increased sensitivity accounts for the greater performance decrease observed for the linear model and Mamba, as illustrated in Table \ref{tab:order_dependency_1}.
To assess the model's ability to capture semantic dependency, we introduce the second assumption.

\textbf{Assumption 2.} Patching enhances the semantic dependency of a sequence \cite{patchtst}.

For example, a single temperature data point is insufficient to illustrate a time pattern, but a continuous set of temperature data over a morning period can reveal valuable insights into the day's climate conditions. Based on Assumption 2, we compare each model's performance variation before and after applying the patching strategy to the input sequence. A more pronounced performance shift post-patching suggests a stronger reliance on semantic dependency.  

The experimental results presented in Table \ref{tab:semantic_learning} underscore a substantial performance gain in the Transformer after applying the patching strategy, significantly outperforming the Linear model under the same conditions. This highlights the Transformer’s superior capability in semantic learning. The Linear model, constrained by its limited expressive power, fails to effectively capture complex semantic dependencies, further justifying the rationale behind our experimental design. Moreover, Mamba, when integrated with the patching strategy, demonstrates even greater improvements over the Linear model, achieving the best overall performance. This result strongly validates Mamba’s effectiveness in modeling semantic dependencies. Notably, as SegRNN has already established the efficacy of the patching strategy, we focus solely on state space models with Mamba in this analysis. 


To summarize, the evaluation results presented above demonstrate that SSMs, including SegRNN and Mamba, uniquely possess the ability to capture both order and semantic dependencies. This capability makes them particularly suitable for LTSF, where maintaining both order and semantic relationships is crucial.

\begin{table}[htbp]
\caption{Results of the linear model, Mamba, and Transformer w/ or w/o patching on ETTm1 dataset.}
\centering
\scalebox{0.6}{
\label{tab:semantic_learning}
\begin{tabular}{c|cc|cc|cc|cc|cc|cc}
\Xhline{1pt}

\multirow{2}{*}{Models} & \multicolumn{4}{c}{Linear Model} & \multicolumn{4}{c}{Mamba} & \multicolumn{4}{c}{Transformer} \\
\cmidrule(lr){2-5} \cmidrule(lr){6-9} \cmidrule(lr){10-13} 
& \multicolumn{2}{c|}{w/o patching}  & \multicolumn{2}{c|}{w/ patching} & \multicolumn{2}{c|}{w/o patching}  & \multicolumn{2}{c|}{w/ patching} & \multicolumn{2}{c|}{w/o patching}  & \multicolumn{2}{c}{w/ patching} \\ 
\cmidrule(lr){2-5} \cmidrule(lr){6-9} \cmidrule(lr){10-13} 
Metrics & \scalebox{1.0}{MSE} & \scalebox{1.0}{MAE} & \scalebox{1.0}{MSE} & \scalebox{1.0}{MAE} & \scalebox{1.0}{MSE} & \scalebox{1.0}{MAE} & \scalebox{1.0}{MSE} & \scalebox{1.0}{MAE} & \scalebox{1.0}{MSE} & \scalebox{1.0}{MAE} & \scalebox{1.0}{MSE} & \scalebox{1.0}{MAE} \\ \midrule
        96 & 0.383& 0.400& 0.366& 0.388& 0.517& 0.508& 0.341& 0.377& 0.643& 0.575& 0.364& 0.394\\
    \midrule
        192 & 0.413& 0.415& 0.400& 0.404& 0.575& 0.546& 0.378& 0.399& 0.805& 0.664& 0.394& 0.404\\
    \midrule
        336 & 0.441& 0.435& 0.429& 0.425& 0.730& 0.634& 0.413& 0.421& 0.882& 0.737& 0.429& 0.430\\ 
    \midrule
        720 & 0.497& 0.469& 0.485& 0.460& 0.873& 0.704& 0.474& 0.465& 0.928& 0.752& 0.468& 0.4600\\ 
    \midrule
        Avg. & 0.434 & 0.469 &   {0.420}&   {0.419}&  0.674& 0.598 &  {0.402} &  {0.416} & 0.815 & 0.682 &  {0.414} &  {0.422}\\     \Xhline{1pt}

\end{tabular}
}


\label{tab:Shuffle}

\end{table}
\subsection{The Pitfalls of Nonlinearity: Overfitting in Deep LTSF Models}
\label{subsection:nonlinear}
\begin{figure}[htbp]
    \centering
    \includegraphics[width=\columnwidth]{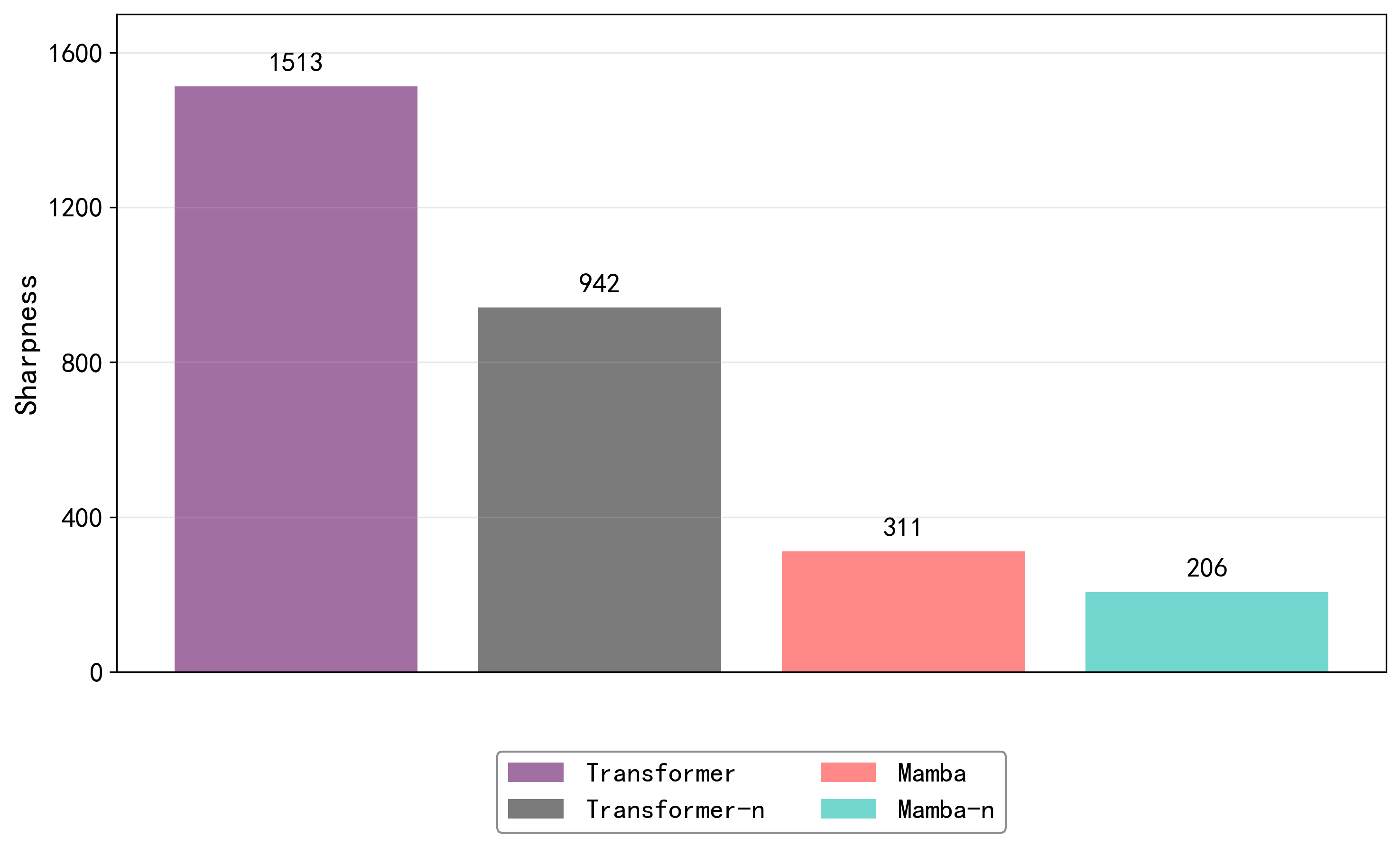}
    \caption{Sharpness at the end of the training. \textit{Transformer-n} refers to removing the activation functions in Transformer, and \textit{Mamba-n} indicates that the Mamba removes activation functions.}
    \label{fig:sharpness_comparison}
\end{figure}
The success of Linear models has prompted researchers to reconsider the utility of more complex architectures, such as Transformers \cite{DLinear}. Initially designed for NLP tasks, these models emphasize semantic dependency over order dependency and incorporate multiple nonlinear activation functions. Consequently, we hypothesize that these nonlinear activation functions may increase the risk of overfitting in time series data, which possess lower information density compared to natural language. 


To test this hypothesis, we carry out ablation studies using nonlinear activation functions across multiple backbone models: Transformer, and Mamba, along with their respective variants that incorporate a patching strategy. As shown in Table \ref{tab:no_linear}, including nonlinear activation functions negatively impacts the model's performance. Interestingly, removing these functions results in notable performance improvements. Figure \ref{fig:sharpness_comparison} further illustrates this trend by showing that sharpness—reflecting the curvature of the loss surface—is typically associated with increased overfitting. Notably, removing the nonlinear activation function leads to a reduction in sharpness, thereby indicating a mitigation of the overfitting problem. This observation implies that while nonlinear activation functions enhance a model's capacity for semantic dependency learning, they may simultaneously impair its ability to generalize from temporal patterns in time series data.

\begin{table}[t]
    \centering
    \small
    \setlength{\tabcolsep}{3pt} 
\caption{Ablation study on nonlinear activation function on ETTm1 dataset. 'Original' means vanilla model, '-n' means removing the nonlinear activation function and 'Patch+' means using patching.}
\label{tab:no_linear}
\resizebox{\linewidth}{!}{ 
    \begin{tabular}{c|cc|cc|cc|cc}
    \Xhline{1pt}
         Models & \multicolumn{2}{c|}{Mamba} & \multicolumn{2}{c|}{Patch+Mamba} & \multicolumn{2}{c|}{Transformer} & \multicolumn{2}{c}{Patch+Transformer} \\ 
         \cmidrule(lr){2-3} \cmidrule(lr){4-5} \cmidrule(lr){6-7} \cmidrule(lr){8-9}
         Metric & MSE & MAE & MSE & MAE & MSE & MAE & MSE & MAE \\
    \midrule
         Original  & 0.674  & 0.598  & 0.395  & 0.419  & 0.815  & 0.682  & 0.414  & 0.422 \\
    \midrule
         Original-n & 0.635  & 0.585  & 0.391  & 0.403  & 0.653  & 0.600  & 0.406  & 0.417 \\
    \midrule
        \textbf{Improvement}  & 5.79\%  & 2.17\%  & 1.01\%  & 1.41\%  & 19.88\%  & 12.02\%  & 1.93\%  & 1.18\% \\ 
    \Xhline{1pt}
    \end{tabular}
}
\end{table}

\begin{table*}[t]
  \caption{ Order dependency analysis. Based on Assumption 1, we compare each model's performance variation before and after randomly shuffling the temporal order of testing data points. O.MSE and O.MAE are evaluated in the original test set. S.MSE and S.MAE are evaluated in the shuffling test set.\emph{Avg.Drop} means the average performance degradation. }
  \label{tab:order_dependency_1}
  \vskip -0.0in
  \vspace{-2pt}
  \renewcommand{\arraystretch}{0.85} 
  \centering
  \resizebox{2.0\columnwidth}{!}{
  \begin{threeparttable}
  \renewcommand{\multirowsetup}{\centering}
  \setlength{\tabcolsep}{1pt}
  \begin{tabular}{cc|cccc|cccc|cccc|cccc|cccc}

    \toprule
    \multicolumn{2}{c}{\multirow{2}{*}{Models}} & 
    \multicolumn{4}{c}{\rotatebox{0}{\scalebox{1.0}{Linear Model}}} &
    \multicolumn{4}{c}{\rotatebox{0}{\scalebox{1.0}{Mamba}}} &
    \multicolumn{4}{c}{\rotatebox{0}{\scalebox{1.0}{SegRNN}}} &
    \multicolumn{4}{c}{\rotatebox{0}{\scalebox{1.0}{Transformer}}} &
    \multicolumn{4}{c}{\rotatebox{0}{\scalebox{1.0}{iTransformer}}} \\
    \cmidrule(lr){3-6} \cmidrule(lr){7-10} \cmidrule(lr){11-14}\cmidrule(lr){15-18}  \cmidrule(lr){19-22}  
    \multicolumn{2}{c}{Metric}  & \scalebox{0.78}{O.MSE} & \scalebox{0.78}{S.MSE}  & \scalebox{0.78}{O.MAE} & \scalebox{0.78}{S.MAE}  & \scalebox{0.78}{O.MSE} & \scalebox{0.78}{S.MSE}  & \scalebox{0.78}{O.MAE} & \scalebox{0.78}{S.MAE} & \scalebox{0.78}{O.MSE} & \scalebox{0.78}{S.MSE}  & \scalebox{0.78}{O.MAE} & \scalebox{0.78}{S.MAE} & \scalebox{0.78}{O.MSE} & \scalebox{0.78}{S.MSE}  & \scalebox{0.78}{O.MAE} & \scalebox{0.78}{S.MAE} & \scalebox{0.78}{O.MSE} & \scalebox{0.78}{S.MSE}  & \scalebox{0.78}{O.MAE} & \scalebox{0.78}{S.MAE}\\
    \toprule
   \multirow{4}{*}{{\rotatebox{90}{\scalebox{0.95}{ETTm1}}}}
    &  \scalebox{0.78}{96}  & 
    {{\scalebox{0.78}{0.383}}} &  {{\scalebox{0.78}{0.988}}} &\scalebox{0.78}{0.400} &\scalebox{0.78}{0.697}&\scalebox{0.78}{0.517} & \scalebox{0.78}{0.922}
    & \scalebox{0.78}{0.508} & \scalebox{0.78}{0.688} & \scalebox{0.78}{0.336} & \scalebox{0.78}{0.982}& \scalebox{0.78}{0.375} & \scalebox{0.78}{0.659}& \scalebox{0.78}{0.643} & \scalebox{0.78}{0.884} &   {{\scalebox{0.78}{0.575}}} &   {{\scalebox{0.78}{0.643}}} & \scalebox{0.78}{0.345} & \scalebox{0.78}{0.892} & \scalebox{0.78}{0.378} & \scalebox{0.78}{0.610}  \\ 
    
    & \scalebox{0.78}{192} &  {{\scalebox{0.78}{0.413}}} &  {{\scalebox{0.78}{0.986}}} &\scalebox{0.78}{0.415} &\scalebox{0.78}{0.697}&  {\scalebox{0.78}{0.575}} &  {\scalebox{0.78}{0.931}}& \scalebox{0.78}{0.546} & \scalebox{0.78}{0.699} & {\scalebox{0.78}{0.376}} &   {{\scalebox{0.78}{0.979}}} & \scalebox{0.78}{0.396} & \scalebox{0.78}{0.662} & \scalebox{0.78}{0.805} & \scalebox{0.78}{1.01} &   {\scalebox{0.78}{0.664}} &   {{\scalebox{0.78}{0.730}}} & \scalebox{0.78}{0.383} & \scalebox{0.78}{0.903} &\scalebox{0.78}{0.395} & \scalebox{0.78}{0.617} \\ 
    
    & \scalebox{0.78}{336} &  {\scalebox{0.78}{0.441}} &  {\scalebox{0.78}{0.987}} &\scalebox{0.78}{0.435} &\scalebox{0.78}{0.698}&  {\scalebox{0.78}{0.730}} &  {\scalebox{0.78}{0.957}} & \scalebox{0.78}{0.634} & \scalebox{0.78}{0.703} & \scalebox{0.78}{0.402} & \scalebox{0.78}{1.070} &  {\scalebox{0.78}{0.415}} &  {\scalebox{0.78}{0.699}} & \scalebox{0.78}{0.882} & \scalebox{0.78}{1.12} &  {\scalebox{0.78}{0.737}} &  {\scalebox{0.78}{0.817}}& \scalebox{0.78}{0.423}  &\scalebox{0.78}{0.923} & \scalebox{0.78}{0.420} & \scalebox{0.78}{0.630} \\ 
    
    & \scalebox{0.78}{720} &  {\scalebox{0.78}{0.497}} &  {\scalebox{0.78}{0.992}}&\scalebox{0.78}{0.469} &\scalebox{0.78}{0.704}& \scalebox{0.78}{0.873} & \scalebox{0.78}{0.973}& \scalebox{0.78}{0.704} & \scalebox{0.78}{0.723} & \scalebox{0.78}{0.458} & \scalebox{0.78}{1.169} &   {{\scalebox{0.78}{0.445}}} &   {{\scalebox{0.78}{0.739}}} & \scalebox{0.78}{0.928} & \scalebox{0.78}{1.12} &   {{\scalebox{0.78}{0.752}}} &   {{\scalebox{0.78}{0.800}}}& \scalebox{0.78}{0.489} & \scalebox{0.78}{0.932} & \scalebox{0.78}{0.456} & \scalebox{0.78}{0.641} \\ 
    \cmidrule(lr){2-22}
    & \scalebox{0.78}{Avg. Drop} &  {\scalebox{0.78}{-}} &  {\scalebox{0.78}{127.97\%}}&\scalebox{0.78}{-} & \scalebox{0.78}{62.55\%} &\scalebox{0.78}{-} & \scalebox{0.78}{40.37\%} &\scalebox{0.78}{-} & \scalebox{0.78}{17.60\%} & \scalebox{0.78}{-} & \scalebox{0.78}{167.18\%} &  {\scalebox{0.78}{-}} &  {\scalebox{0.78}{69.16\%}} & \scalebox{0.78}{-} & \scalebox{0.78}{22.40\%} &  {\scalebox{0.78}{-}} &  {\scalebox{0.78}{6.55\%}} & \scalebox{0.78}{-} & \scalebox{0.78}{122.56\%} & \scalebox{0.78}{-} & \scalebox{0.78}{51.5\%} \\ 
        \midrule
    
    \multirow{4}{*}{\rotatebox{90}{{\scalebox{0.95}{Exchange}}}}
    &  \scalebox{0.78}{96} 
    & {\scalebox{0.78}{0.0832}} &  {\scalebox{0.78}{0.210}}
&\scalebox{0.78}{0.201} & \scalebox{0.78}{0.332}
    & {\scalebox{0.78}{1.260}} &  {\scalebox{0.78}{1.401}}
    & {\scalebox{0.78}{0.915}} & {\scalebox{0.78}{0.943}} & \scalebox{0.78}{0.083} & \scalebox{0.78}{0.252} & {\scalebox{0.78}{0.200}} & {\scalebox{0.78}{0.356}}  & \scalebox{0.78}{0.730} & \scalebox{0.78}{0.738} & {\scalebox{0.78}{0.782}} & {\scalebox{0.78}{0.722}}& \scalebox{0.78}{0.0869} & \scalebox{0.78}{0.242} & \scalebox{0.78}{0.207} & \scalebox{0.78}{0.358}  \\

    &  \scalebox{0.78}{192}
    & {\scalebox{0.78}{0.179}} &  {\scalebox{0.78}{0.325}}
    &\scalebox{0.78}{0.299} & \scalebox{0.78}{0.414}
   & {\scalebox{0.78}{1.398}} &  {\scalebox{0.78}{1.626}}
    & \scalebox{0.78}{1.040} & {\scalebox{0.78}{1.060}} & \scalebox{0.78}{0.176} & \scalebox{0.78}{0.374} &  {\scalebox{0.78}{0.296}} & {\scalebox{0.78}{0.436}} & \scalebox{0.78}{1.304} & \scalebox{0.78}{1.284} &  {\scalebox{0.78}{0.913}} & {\scalebox{0.78}{0.949}}& \scalebox{0.78}{0.179} & \scalebox{0.78}{0.374} & \scalebox{0.78}{0.301} & \scalebox{0.78}{0.450}\\

    &  \scalebox{0.78}{336} 
    &\scalebox{0.78}{0.338} &  {\scalebox{0.78}{0.521}}
     &\scalebox{0.78}{0.418} & \scalebox{0.78}{0.534}
    &\scalebox{0.78}{1.835} & \scalebox{0.78}{1.921}
    & \scalebox{0.78}{1.111} & {\scalebox{0.78}{1.141}} & \scalebox{0.78}{0.332} & \scalebox{0.78}{0.531}&  {\scalebox{0.78}{0.416}} &  {\scalebox{0.78}{0.537}}& \scalebox{0.78}{1.860} & \scalebox{0.78}{1.862}&  {\scalebox{0.78}{1.090}} &  {\scalebox{0.78}{1.085}} & \scalebox{0.78}{0.331} & \scalebox{0.78}{0.555} & \scalebox{0.78}{0.417} & \scalebox{0.78}{0.557} \\ 
    &  \scalebox{0.78}{720} 
    & {\scalebox{0.78}{0.903}} &  {\scalebox{0.78}{1.167}}
    &\scalebox{0.78}{0.714} & \scalebox{0.78}{0.822}
    &\scalebox{0.78}{3.940} & \scalebox{0.78}{4.023}
    &  {\scalebox{0.78}{1.687}} &  {\scalebox{0.78}{1.697}}&  {\scalebox{0.78}{0.935}} & \scalebox{0.78}{1.190} & \scalebox{0.78}{0.736} & \scalebox{0.78}{0.836}&   {\scalebox{0.78}{3.860}} & \scalebox{0.78}{3.865} & \scalebox{0.78}{1.684} & \scalebox{0.78}{1.685}& \scalebox{0.78}{0.856} & \scalebox{0.78}{1.202} & \scalebox{0.78}{0.698} & \scalebox{0.78}{0.841}   
    \\ 
    \cmidrule(lr){2-22}
    &  \scalebox{0.78}{Avg. Drop}     & {\scalebox{0.78}{-}} &  {\scalebox{0.78}{47.89\%}}
    &\scalebox{0.78}{-} & \scalebox{0.78}{28.80\%}
    &\scalebox{0.78}{-} & \scalebox{0.78}{6.38\%}
    & {\scalebox{0.78}{-}} &  {\scalebox{0.78}{1.85\%}} & \scalebox{0.78}{-} & \scalebox{0.78}{53.80\%} & \scalebox{0.78}{-} & {\scalebox{0.78}{31.37\%}} & \scalebox{0.78}{-} & \scalebox{0.78}{-0.06\%} & \scalebox{0.78}{-} & {\scalebox{0.78}{-0.63\%}} & \scalebox{0.78}{-} & \scalebox{0.78}{63.33\%} & \scalebox{0.78}{-} & \scalebox{0.78}{35.89\%}\\

    \midrule
    
    \multirow{4}{*}{\rotatebox{90}{\scalebox{0.95}{Traffic}}} 
    & \scalebox{0.78}{96} 
    &  {\scalebox{0.78}{0.656}} &  {\scalebox{0.78}{1.679}} &\scalebox{0.78}{0.403} &\scalebox{0.78}{0.882}& {\scalebox{0.78}{0.634}} & {\scalebox{0.78}{1.859}} &  {\scalebox{0.78}{0.363}} &  {\scalebox{0.78}{0.967}} & \scalebox{0.78}{0.651} & \scalebox{0.78}{1.634} & {\scalebox{0.78}{0.327}} & \scalebox{0.78}{0.850}  & \scalebox{0.78}{0.788} & \scalebox{0.78}{1.846} & {\scalebox{0.78}{0.467}} & \scalebox{0.78}{0.918}& \scalebox{0.78}{0.396} & {\scalebox{0.78}{1.945}} & \scalebox{0.78}{0.271} & \scalebox{0.78}{0.976} \\ 
    & \scalebox{0.78}{192} 
    &  {\scalebox{0.78}{0.609}} &  {\scalebox{0.78}{1.665}}&\scalebox{0.78}{0.382} &\scalebox{0.78}{0.876}& {\scalebox{0.78}{0.637}} & {\scalebox{0.78}{2.387}}&  {{\scalebox{0.78}{0.375}}} &  {{\scalebox{0.78}{1.120}}} & \scalebox{0.78}{0.657} & \scalebox{0.78}{1.633} & {\scalebox{0.78}{0.332}} & \scalebox{0.78}{0.851} & \scalebox{0.78}{0.850} & \scalebox{0.78}{1.715} & {\scalebox{0.78}{0.483}} & \scalebox{0.78}{0.878} & \scalebox{0.78}{0.416} & {\scalebox{0.78}{1.751}} & \scalebox{0.78}{0.279} & \scalebox{0.78}{0.916}  \\ 
    & \scalebox{0.78}{336} 
    &  {\scalebox{0.78}{0.615}} &  {\scalebox{0.78}{1.671}}&\scalebox{0.78}{0.385} &\scalebox{0.78}{0.880}& {\scalebox{0.78}{0.674}} & {\scalebox{0.78}{2.428}}&  {\scalebox{0.78}{0.386}} &  {\scalebox{0.78}{1.127}} & \scalebox{0.78}{0.678} & \scalebox{0.78}{1.636} & {\scalebox{0.78}{0.341}} & {\scalebox{0.78}{0.853}}  & \scalebox{0.78}{0.836} & \scalebox{0.78}{1.676} & {\scalebox{0.78}{0.480}} & {\scalebox{0.78}{0.861}}& \scalebox{0.78}{0.430} & \scalebox{0.78}{1.801}  & \scalebox{0.78}{0.287} & \scalebox{0.78}{0.938} \\

    & \scalebox{0.78}{720} 
    &  {\scalebox{0.78}{0.656}} &  {\scalebox{0.78}{1.697}} &\scalebox{0.78}{0.405} &\scalebox{0.78}{0.882}& {\scalebox{0.78}{0.732}} & {\scalebox{0.78}{1.751}} &  {\scalebox{0.78}{0.414}} &  {\scalebox{0.78}{0.916}} & \scalebox{0.78}{0.676} & \scalebox{0.78}{1.779} & {\scalebox{0.78}{0.327}} & {\scalebox{0.78}{0.900}} & \scalebox{0.78}{0.859} & \scalebox{0.78}{1.746} & {\scalebox{0.78}{0.490}} & {\scalebox{0.78}{0.880}} & \scalebox{0.78}{0.559} & \scalebox{0.78}{1.978}  & \scalebox{0.78}{0.375} & \scalebox{0.78}{0.971}  \\ 
    \cmidrule(lr){2-22}
    & \scalebox{0.78}{Avg. Drop} & {\scalebox{0.78}{-}} &  {\scalebox{0.78}{164.67\%}}
    &\scalebox{0.78}{-} 
    & \scalebox{0.78}{123.49\%}
    &\scalebox{0.78}{-} 
    & \scalebox{0.78}{214.72\%}&  {\scalebox{0.78}{-}} &  {\scalebox{0.78}{168.53\%}} & \scalebox{0.78}{-} 
    & \scalebox{0.78}{149.33\%} & {\scalebox{0.78}{-}} & {\scalebox{0.78}{160.29\%}}&   \scalebox{0.78}{-} & \scalebox{0.78}{109.51\%} & {\scalebox{0.78}{-}} & {\scalebox{0.78}{84.22\%}}&\scalebox{0.78}{-} & {\scalebox{0.78}{315.05\%}} & \scalebox{0.78}{-} & \scalebox{0.78}{213.61\%}  \\ 
    \bottomrule
  \end{tabular}
  \end{threeparttable}
}
\end{table*}
    
    

Our analysis further reveals that the adverse effects of nonlinear activation functions are most pronounced in Transformer-based architectures, with Mamba trailing. Interestingly, this order is inversely correlated with the model's ability to capture order dependency. This implies that a model's capability to grasp order dependency can partially mitigate the overfitting caused by nonlinear activation functions, thereby enhancing its potential to generalize temporal patterns.

\section{SDE: Simplified and Disentangled Dependency Encoding}

As aforementioned, adapting SSMs to LTSF presents two non-trivial challenges. First, existing SSMs tend to overly rely on semantic dependencies, which may obscure the intrinsic temporal dynamics. Second, effectively capturing cross-variate dependencies without compromising the distinct characteristics of individual variates. To address these challenges, we propose \textbf{SDE} (\textbf{S}implified and \textbf{D}isentangled Dependency \textbf{E}ncoding), which consists of two core components, each designed to mitigate one of these issues.

The first component, Simplification Mechanism, is motivated by our analysis of Mamba's training dynamics in Section \ref{subsection:nonlinear}, where we identify an overfitting issue stemming from excessive nonlinear activation. Inspired by this observation, we introduce a structured simplification strategy that removes unnecessary nonlinear operations traditionally used for learning complex representations.

The second component is a theoretically grounded disentangled dependency encoding strategy, which effectively disentangles temporal (order and semantic dependencies) and cross-variate dependencies. Unlike conventional approaches that encode these dependencies sequentially—often leading to entangled representations that dilute meaningful relationships—our framework explicitly models them in parallel. This disentanglement ensures that cross-variate interactions are captured effectively while preserving the distinct temporal dynamics of each variate.

\begin{figure*}[t]
  \centering
  \begin{minipage}[b]{0.72\linewidth}
    \centering
    \includegraphics[width=\linewidth]{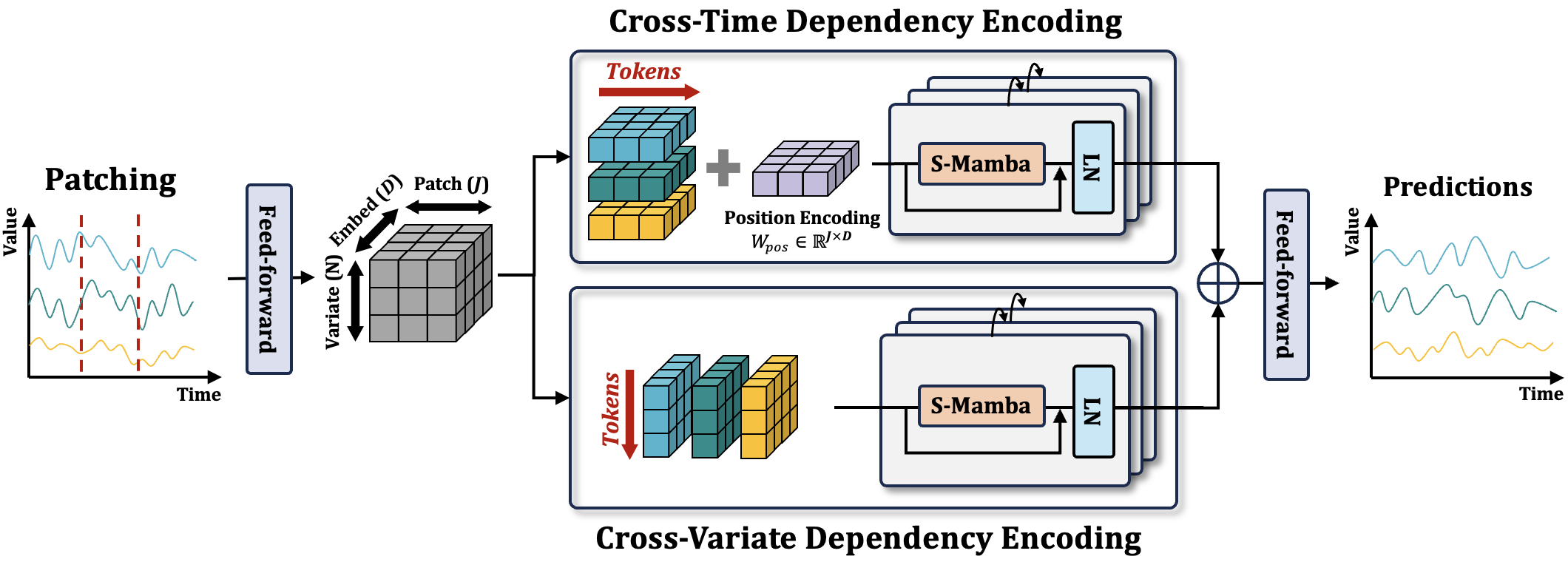}
    \caption{The overall framework of \model.}
    \label{fig:enter-label}
  \end{minipage}
  \begin{minipage}[b]{0.25\linewidth}
    \centering
    \includegraphics[width=\linewidth]{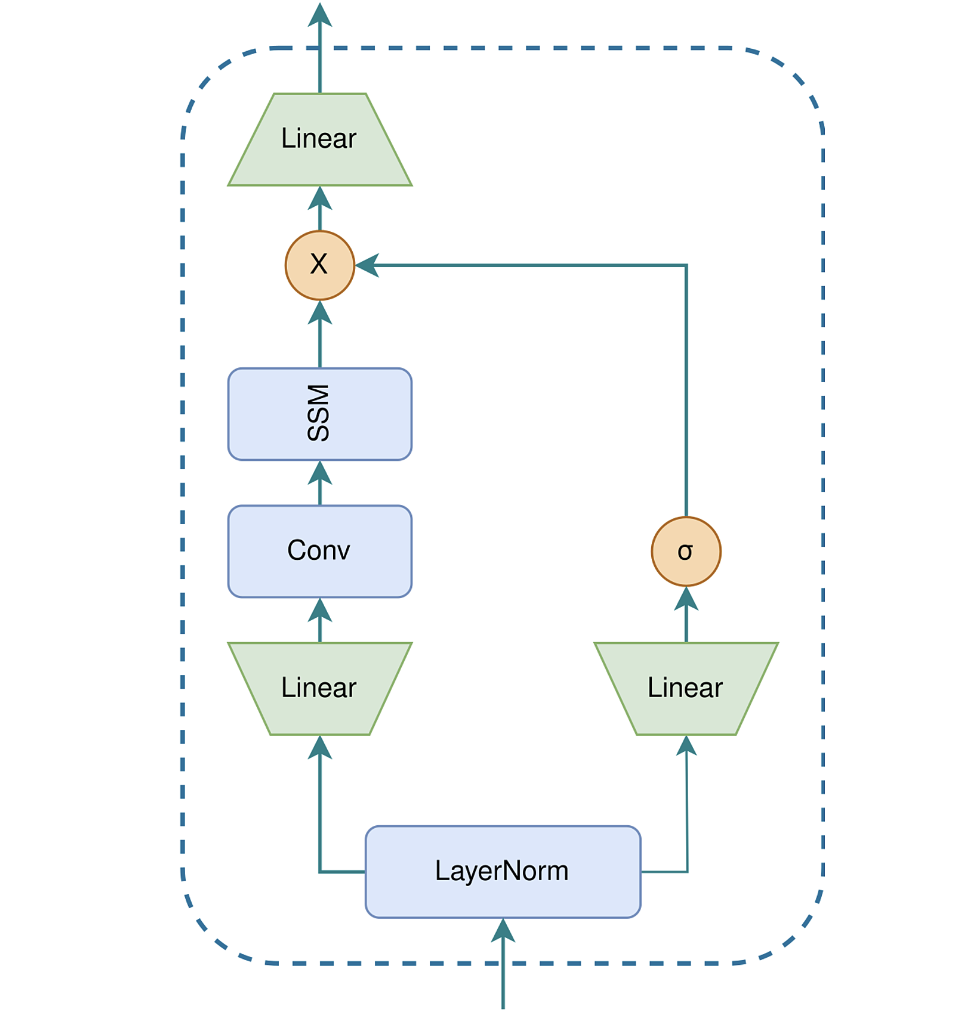} 
    \caption{A S-Mamba block.}  
    \label{fig:2}
  \end{minipage}
\end{figure*}

To facilitate a clear and structured introduction to our approach, \textbf{we take Mamba as a representative example} to illustrate the details and procedural steps of the SDE framework. The resulting architecture, \model, is illustrated in Figure \ref{fig:enter-label}.

\subsection{Simplification Mechanism for Mamba}
Although attention mechanism excel in learning semantic dependency, its inherent permutation invariance constrains its ability to accurately capture order dependency, despite the use of psitional encoding \cite{DLinear}. Compared to Transformer, Mamba's strength in LTSF lies in its use of selective SSM that is similar to self-attention but processes time series recursively. This approach enables Mamba to capture order dependency and semantic dependency simultaneously, as demonstrated in Section \ref{subsection:mamba}. Formally, selective SSM can be described as: 
\begin{align*}
    & \bm{h}_t = \overline{\mathbf{A}}\bm{h}_{t-1} + \overline{\mathbf{B}} \bm{x}_t, \ \bm{y}_t = \mathbf{C}\bm{h}_t.
\end{align*}
where $\overline{\mathbf{A}}\in\mathbb{R}^{N \times N}$, and $\overline{\mathbf{B}},\mathbf{C}\in\mathbb{R}^{N \times D}$ are learnable parameters that map input sequence $\bm{x}_t\in\mathbb{R}^{D}$ to output sequence $\bm{y}_t\in\mathbb{R}^{D}$ through an hidden state $\bm{h}_t\in\mathbb{R}^N$. In particular, $\overline{\mathbf{A}}$ and $\overline{\mathbf{B}}$ are the discretized forms of ${\mathbf{A}}$ and ${\mathbf{B}}$ using $\boldsymbol{\Delta}$ for seamless intergration deep learning. The discretization functions for the input time series are defined as: 
\begin{align*}
    &\overline{\mathbf{A}}=\exp{(\Delta\mathbf{A})},\ \overline{\mathbf{B}}=(\boldsymbol{\Delta}\mathbf{A})^{-1}(\exp{(\boldsymbol{\Delta}\mathbf{A})}-\mathbf{I})\cdot\boldsymbol{\Delta}\mathbf{B}.
\end{align*}
Based on discrete SSM that has been theoretically proven to capture long-range order dependency \cite{gu2021combining}, Mamba introduces a selection mechanism that makes $\mathbf{B}$, $\mathbf{C}$ and $\boldsymbol{\Delta}$ data dependent, allowing it to capture semantic dependency as well:
\[
    \mathbf{B} = \text{Linear}_N(\bm{x}_t), \quad \mathbf{C} = \text{Linear}_N(\bm{x}_t), \quad \boldsymbol{\Delta} = \text{softplus}(\text{Linear}_N(\bm{x}_t)).
\]

In addition to the slective SSM, each Mamba layer consists of two branches. The left branch show in Figure \ref{fig:2} is expressed as:
\begin{align*}
    \bm{x}_t' &= \text{SelectiveSSM}(\sigma (\text{Conv1D} (\text{Linear} (\bm{x}_t)))).
\end{align*}
This branch is designed to efficiently capture both order and semantic dependencies. Building on our analysis in Section \ref{subsection:nonlinear} and the proposed Simplification Mechanism, we eliminate the nonlinear activation function between the Conv1D and SSM layers in Mamba, leading to the development of S-Mamba. This design choice effectively mitigates the overfitting issue, as 1D convolution, similar to a linear layer, is sufficiently expressive to capture complex time series representations without introducing unnecessary nonlinearity.

The remaining part is the implementation of a gating mechanism and residual operation, which can be written as:
\begin{align*}
    \bm{y} &= \text{LayerNorm} (  \bm{x}_t'\otimes (\sigma (\text{Linear} (\bm{x}_t))) + \bm{x}_t).
\end{align*}
Notice that we retain the nonlinear activation within the gating mechanism (right in Figure~\ref{fig:2}) to ensure the stability and robustness of the learning process \cite{gu2023mamba,chung2014empirical}.

\subsection{Disentangled Dependency Encoding }
Previous CD approaches encode cross-time dependency (order and semantic dependencies) and cross-variate dependency sequentially. However, this paradigm can introduce less relevant cross-time dependency from other variates, negatively impacting the informativeness of the embedding for the target variate. To address this, we propose a theoretically sound disentangled encoding strategy that processes cross-time and cross-variate dependencies in parallel. By aggregating information from both dimensions and mapping them into a unified space, our strategy ensures that these dependencies are fully leveraged while minimizing the negative interference between them.

\textbf{Dependency-specific Embedding.}
The intricate interactions within multivariate time series make disentangling cross-time and cross-variate dependencies a complex challenge. Analyzing the inputs of Figure \ref{fig:enter-label}, we observe that different variates exhibit various periodicities and value ranges but show evident regularity in local changes. This insight motivated us to consider dependencies from the subseries level. Consequently, we divide the $i$-th variate time series $\mathbf{c}^i$ into several patches $\mathbf{C}_p^i\in\mathbb{R}^{J\times P}$, where $J$ is the number of patches and $P$ refers to the length of each patch. Subsequently, we use a linear projection ${\mathbf{W}}_{p}\in\mathbb{R}^{P\times D}$ to map each patch into the $D$ dimensional embedding space as $\mathbf{E}^i\in\mathbb{R}^{J\times D}.$
Thereafter, we concatenate the embeddings of the patch into a tensor $\mathbf{E}\in\mathbb{R}^{N\times J\times D}$, where $N$ denotes the number of variates. To separately capture the cross-time and cross-variate dependencies, we disentangle the tensor along the temporal dimension as $\mathbf{E}_{time}^i\in\mathbb{R}^{J\times D}$ and along the variate dimension as $\mathbf{E}_{var}^i\in\mathbb{R}^{N\times D}$.

\textbf{Cross-time Dependency Modeling.}
To accurately capture cross-time dependency, we independently model each variate. First, we introduce the learnable additive position encoding $\mathbf{W}_{\mathrm{pos}}\in\mathbb{R}^{J\times D}$ to enhance the order information of the patches, modifying the embedding as $\mathbf{E}_{\text{time}}^i \leftarrow \mathbf{E}_{\text{time}}^i+\mathbf{W}_{\mathrm{pos}}$, where $\mathbf{E}_{\text{time}}^i\in\mathbb{R}^{J\times D}$. Subsequently, we process each channel independently through a single $\mathcal{F}$ block, equipped with residual connections and Layer Normalization (LN) \cite{layernorm}, formulated as $\mathbf{E}_{\text{time}}^i = \text{LN}(\mathcal{F}(\mathbf{E}_{\text{time}}^i) + \mathbf{E}_{\text{time}}^i)$. Here, $\mathcal{F}$ represents a generic sequence encoder, which can be instantiated as a Transformer, Mamba, or RNN, depending on the specific modeling needs. In our present implementation, we utilize S-Mamba as the encoder. This process effectively extracts both order and semantic dependencies for each variate, producing a patch representation $\mathbf{E}_{\text{time}}\in\mathbb{R}^{N\times J\times D}$ enriched with temporal information.


\textbf{Cross-variate Dependency Modeling.}
Existing approaches to modeling cross-variate dependency often entangle temporal and variate information, thereby blurring the distinction between individual channels \cite{liu2023itransformer}. This entanglement can degrade model performance by introducing spurious correlations and reducing the informativeness of learned representations. To overcome this limitation, we propose a unified cross-variate dependency encoder within our disentangled dependency encoding strategy, designed to disentangle these dependencies and preserve the structural integrity of each variate. In particular, this encoder is a modular component that can be instantiated with various architectures, such as Mamba, Transformers, or RNNs, allowing for adaptability to different modeling paradigms.

In our current instantiation, we employ S-Mamba as the cross-variate dependency encoder. SSMs, including Mamba and RNNs, inherently process inputs sequentially, which is not naturally suited for capturing unordered cross-variate dependency. To address this, we adapt S-Mamba to operate on an unordered set of variates by simply applying it to a random permutation of the variate order.

Formally, given an input sequence $\mathbf{E}_{\text{var}}^i\in\mathbb{R}^{N\times D}$, we apply S-Mamba over a randomly permuted variate order, generating the transformed representation $\mathbf{E}_{\text{fo}}^i$. To effectively integrate cross-variate dependency while preserving the distinct characteristics of each variate, we independently process patches corresponding to the same time step along the variate dimension, formulated as $\mathbf{E}_{\text{var}}^i = \text{LN}( \mathbf{E}_{\text{fo}}^i + \mathbf{E}_{\text{var}}^i)$. This encoding strategy produces patch representations $\mathbf{E}_{\text{var}}\in\mathbb{R}^{J\times N\times D}$, effectively capturing complex cross-variate dependencies while maintaining a disentangled and structured representation.

\textbf{Unified Representation for Forecasting.}
After disentangling and separately capturing cross-time and cross-variate dependencies, we concatenate the corresponding representations and utilize a Feed-Forward Network (FFN) to aggregate and map them into a unified space: $\mathbf{E}_{o}=\text{FFN}(\mathbf{E}_{time} || \mathbf{E}_{var}) \in\mathbb{R}^{N\times J\times D}$.
In this way, we obtain patch representations that capture three key dependencies without introducing harmful interference. Finally, we use a flatten layer with a linear head to map these patch representations $\mathbb{E}_{o}$ to the final prediction result $\mathbf{Z}\in\mathbb{R}^{N\times S}$.

\textbf{Theoretical Discussions.}
Finally, we theoretically analyze the sufficient expressiveness of our disentanglement strategy compared to LSTF models that alternatively encode cross-time and cross-variate dependencies. To begin with, let $\Phi:\mathbb{R}^{J\times D} \rightarrow \mathbb{R}^{J\times D}$ and $\Psi:\mathbb{R}^{N\times D} \rightarrow \mathbb{R}^{N\times D}$ denote a cross-time encoder and a cross-variate encoder, respectively. With a little abuse of notation, a broadcasting mechanism will be applied when encoding 3D tensor inputs. Then our disentangled model can be denoted as $\mathcal{F}_{d}(\cdot) = \text{FFN}(\Phi(\cdot) || \Psi(\cdot)),$ which encodes the two types of dependencies in parallel. Based on Theorem 3.5 in \cite{time_then_graph2022icml} and the evidence from \cite{liu2023itransformer}, we focus on comparing our model with \emph{time-then-variate} LSTF models, which can be defined as $\mathcal{F}_{ttv}(\cdot) = \text{FFN}(\Psi(\Phi(\cdot))).$ 
Without loss of generality, we only consider the single-layer setting. 
We further assume that the cross-time encoder $\Phi(\cdot)$ 
can be decomposed into $\Phi(\cdot) = \phi(\mathcal{O}(\cdot), \mathcal{S}(\cdot)) + \mathcal{Z}(\cdot),$ where $\mathcal{O}(\cdot)$ and $\mathcal{S}(\cdot)$ extract order and semantic dependency respectively, $\mathcal{Z}(\cdot)$ is the noisy component, and $\phi(\cdot)$ is a mapping. Let $\mathbf{c}^i := \mathbf{c}^i_{T+1:T+S}$ for simplicity. Our main theorem is as follows.

\begin{myThm}
For any variate $i$, $1 \leq i \leq N,$ if (1) $I(\mathbf{c}^i;\mathcal{Z}(\mathbf{E})^j |$ $\mathcal{O}(\mathbf{E})^j, \mathcal{S}(\mathbf{E})^j)=0, \forall 1\le j\le N$ and (2) the dependencies of different variates satisfy $I(\mathbf{c}^i; \mathcal{S}(\mathbf{E})^{-i} | \Phi(\mathbf{E})^i, \mathcal{O}(\mathbf{E})^{-i}) \leq \epsilon$ for some $\epsilon \ge 0,$ where $\mathcal{O}(\mathbf{E})^{-i}$($\mathcal{S}(\mathbf{E})^{-i}$) denotes the order(semantic) dependencies of variates except from the $i$-th one, then the informativeness of the representations output by the disentangled model and the time-then-variate model satisfy $H(\mathbf{c}^i | \mathcal{F}_d(\mathbf{E})^i) \leq H(\mathbf{c}^i | \mathcal{F}_{ttv}(\mathbf{E})^i) + \epsilon.$ 

\end{myThm}
Please refer to \appref{app:thm} for detailed proof. Intuitively, given the assumption that cross-time encoding cannot extract more informative semantic dependencies from the embeddings of other variates for forecasting the future of the target variate, our disentanglement strategy can extract sufficient dependencies while reducing interference between cross-time and cross-variate encoding processes. More discussions on the rationality of the assumption can be found in \appref{app:thm_assumption}
\section{Experiments}
We comprehensively evaluate the proposed SDE framework's performance and efficiency in LTSF and analyze the effectiveness of each component. Meanwhile, we extend our disentangled strategy to other models, validating the proposed framework's universality.
We briefly list the datasets and baselines below. The implementation details are provided in Appendix \ref{appendix:exper_details}.

\textbf{Datasets.} We conduct experiments on nine real-world datasets following \cite{liu2023itransformer}: (1) ECL, (2) ETTh1, (3) ETTh2, (4) ETTm1, (5) ETTm2, (6) Exchange, (7) Traffic, (8) Weather, (9) Solar-Energy.

\textbf{Baselines.} We carefully use 14 popular LTSF forecasting models as our baselines and cite their performance from \cite{liu2023itransformer} if applicable. Our baselines include (1)~\emph{Pretrained Language model-based model:} FTP (\cite{GPT4TS}); (2)~\emph{Transformer-based models:} Autoformer \cite{wu2021autoformer}, FEDformer \cite{fedformer}, Stationary \cite{Non_stationary}, Crossformer \cite{crossformer}, PatchTST \cite{patchtst}, iTransformer \cite{liu2023itransformer}, CARD \cite{wang2024card}; (3)~\emph{Linear-based models:} DLinear \cite{DLinear}, TiDE \cite{tide}, RLinear \cite{RLinear}; (5)~\emph{RNN-based model:} SegRNN \cite{lin2023segrnn}; and (4)~\emph{TCN-based models:} SCINet \cite{SCINet}, TimesNet \cite{wu2022timesnet}.

\subsection{Experiment Details}
\label{appendix:exper_details}
All experiments are implemented using PyTorch on a single NVIDIA A40 GPU. We optimize the model using ADAM with an initial learning rate in the set $\{10^{-3}, 5\times 10^{-4}, 10^{-4}\}$ and L2 loss. The batch size is uniformly set to $\{32-128\}$, depending on the maximum GPU capacity. To ensure that all models can be fitted effectively, the number of training epochs is fixed at $10$ with an early stopping mechanism. By default, our \model encodes both the time and variate dimensions using a single-layer \model encoding, with a patch length of $16$ and a stride of $8$.

The baseline models we use for comparison are implemented based on the Time Series Library (\cite{wu2022timesnet}) repository, and they are run with the hyperparameter settings they recommend. For CARD, we download the model from \text{https://github.com/wxie9/CARD/tree/main}, but utilize the same training loss as the other models (i.e., the Time Series Library GitHub repository) recommend for a fair comparison. Additionally, the full results of the predictions come from the outcomes reported by the iTransformer (\cite{liu2023itransformer}), ensuring fairness.

\subsection{Long-Term Time Series Forecasting}
\label{LTSF_results}
Table \ref{tab:forecasting_results} presents the multivariate long-term forecasting results. In general, \model outperforms all baseline methods. Specifically, for the ETT dataset, which exhibits weak cross-variate dependency, models using the CD strategy, such as iTransformer and Crossformer, underperform the models that use the CI strategy (e.g., PatchTST). However, \model, which introduces cross-variate dependency through a disentangled encoding strategy, demonstrates superior performance over all models that use the CI strategy. For datasets with significant cross-variate dependency, such as Weather, ECL, and Traffic, \model also performs comparable to or superior to the SOTA iTransformer. 
Furthermore, the table highlights a remarkable performance boost of SDE-SegRNN over its predecessor, SegRNN, reinforcing the efficacy of our SDE framework in improving state-space models (SSMs). For instance, on the Solar-Energy dataset, SDE-SegRNN reduces the MSE from 0.239 (SegRNN) to 0.225, while on Traffic, it improves the MSE from 0.670 (SegRNN) to 0.622, confirming its ability to enhance predictive accuracy across diverse datasets. These results validate the effectiveness of our proposed framework in advancing multivariate long-term forecasting.
\begin{table*}[htbp]
  \caption{\update{Multivariate forecasting results} with prediction lengths $S\in\{96, 192, 336, 720\}$ and fixed lookback length $T=96$. Results are averaged from all prediction lengths. Full results are listed in Table \ref{tab:full_forecasting_results}.}
  \label{tab:forecasting_results}
  \renewcommand{\arraystretch}{0.85} 
  \centering
  \resizebox{2.1\columnwidth}{!}{
  \begin{threeparttable}
  \begin{small}
  \renewcommand{\multirowsetup}{\centering}
  \setlength{\tabcolsep}{1.45pt}
  \begin{tabular}{c|cc|cc|cc|cc|cc|cc|cc|cc|cc|cc|cc|cc|cc|cc|cc|cc}
    \toprule
    {\multirow{2}{*}{Models}} & 
    \multicolumn{2}{c}{\rotatebox{0}{\scalebox{0.75}{\textbf{\model}}}} &
     \multicolumn{2}{c}{\rotatebox{0}
    {\scalebox{0.8}{\update{SDE-SegRNN}}}}&
     \multicolumn{2}{c}{\rotatebox{0}
    {\scalebox{0.8}{\update{SegRNN}}}}&
     \multicolumn{2}{c}{\rotatebox{0}
    {\scalebox{0.8}{\update{CARD}}}} &
    \multicolumn{2}{c}{\rotatebox{0}
    {\scalebox{0.8}{\update{FTP}}}} &
    \multicolumn{2}{c}{\rotatebox{0}
    {\scalebox{0.8}{\update{iTransformer}}}} &
    \multicolumn{2}{c}{\rotatebox{0}
    {\scalebox{0.8}{\update{RLinear}}}} &
    \multicolumn{2}{c}{\rotatebox{0}{\scalebox{0.8}{PatchTST}}} &
    \multicolumn{2}{c}{\rotatebox{0}{\scalebox{0.8}{Crossformer}}} &
    \multicolumn{2}{c}{\rotatebox{0}{\scalebox{0.8}{TiDE}}} &
    \multicolumn{2}{c}{\rotatebox{0}{\scalebox{0.8}{{TimesNet}}}} &
    \multicolumn{2}{c}{\rotatebox{0}{\scalebox{0.8}{DLinear}}} &
    \multicolumn{2}{c}{\rotatebox{0}{\scalebox{0.8}{SCINet}}} &
    \multicolumn{2}{c}{\rotatebox{0}{\scalebox{0.8}{FEDformer}}} &
    \multicolumn{2}{c}{\rotatebox{0}{\scalebox{0.8}{Stationary}}} &
    \multicolumn{2}{c}{\rotatebox{0}{\scalebox{0.8}{Autoformer}}}  \\
     &
   \multicolumn{2}{c}{\scalebox{0.8}{\textbf{(Ours)}}} &
   \multicolumn{2}{c}{\scalebox{0.8}{\textbf{(Ours)}}}&
   \multicolumn{2}{c}{\scalebox{0.8}{\citeyearpar{lin2023segrnn}}}&
   \multicolumn{2}{c}{\scalebox{0.8}{\citeyearpar{wang2024card}}} &
      \multicolumn{2}{c}{\scalebox{0.8}{\citeyearpar{GPT4TS}}} &
     \multicolumn{2}{c}{\scalebox{0.8}{\citeyearpar{liu2023itransformer}}} & 
    \multicolumn{2}{c}{\scalebox{0.8}{\citeyearpar{RLinear}}} & 
    \multicolumn{2}{c}{\scalebox{0.8}{\citeyearpar{patchtst}}} & 
    \multicolumn{2}{c}{\scalebox{0.8}{\citeyearpar{crossformer}}} & 
    \multicolumn{2}{c}{\scalebox{0.8}{\citeyearpar{tide}}} & 
    \multicolumn{2}{c}{\scalebox{0.8}{\citeyearpar{wu2022timesnet}}} & 
    \multicolumn{2}{c}{\scalebox{0.8}{\citeyearpar{DLinear}}} &
    \multicolumn{2}{c}{\scalebox{0.8}{\citeyearpar{SCINet}}} & 
    \multicolumn{2}{c}{\scalebox{0.8}{\citeyearpar{fedformer}}} &
    \multicolumn{2}{c}{\scalebox{0.8}{\citeyearpar{Non_stationary}}} &
    \multicolumn{2}{c}{\scalebox{0.8}{\citeyearpar{wu2021autoformer}}} \\
    \cmidrule(lr){2-3} \cmidrule(lr){4-5}\cmidrule(lr){6-7} \cmidrule(lr){8-9}\cmidrule(lr){10-11}\cmidrule(lr){12-13} \cmidrule(lr){14-15} \cmidrule(lr){16-17} \cmidrule(lr){18-19} \cmidrule(lr){20-21} \cmidrule(lr){22-23} \cmidrule(lr){24-25} \cmidrule(lr){26-27}  \cmidrule(lr){28-29}
    {Metric}  & \scalebox{0.8}{MSE} & \scalebox{0.8}{MAE}  & \scalebox{0.8}{MSE} & \scalebox{0.8}{MAE}  & \scalebox{0.8}{MSE} & \scalebox{0.8}{MAE}  & \scalebox{0.8}{MSE} & \scalebox{0.8}{MAE}  & \scalebox{0.8}{MSE} & \scalebox{0.8}{MAE}  & \scalebox{0.8}{MSE} & \scalebox{0.8}{MAE} & \scalebox{0.8}{MSE} & \scalebox{0.8}{MAE} & \scalebox{0.8}{MSE} & \scalebox{0.8}{MAE} & \scalebox{0.8}{MSE} & \scalebox{0.8}{MAE} & \scalebox{0.8}{MSE} & \scalebox{0.8}{MAE} & \scalebox{0.8}{MSE} & \scalebox{0.8}{MAE} & \scalebox{0.8}{MSE} & \scalebox{0.8}{MAE}& \scalebox{0.8}{MSE} & \scalebox{0.8}{MAE}&\scalebox{0.8}{MSE} & \scalebox{0.8}{MAE}& \scalebox{0.8}{MSE} & \scalebox{0.8}{MAE}&\scalebox{0.8}{MSE} & \scalebox{0.8}{MAE}\\
    \toprule
    \scalebox{0.95}{ECL} 
      & 
    \boldres{\scalebox{0.8}{0.172}} & \boldres{\scalebox{0.8}{0.268}}& 
    \secondres{\scalebox{0.8}{0.173}} & \scalebox{0.8}{0.271}
     & 
     \scalebox{0.8}{0.189} & \scalebox{0.8}{0.282}
     & 
     \scalebox{0.8}{0.204} & \scalebox{0.8}{0.291}
     & 
    \scalebox{0.8}{0.210} & \scalebox{0.8}{0.291}
    & 
    {\scalebox{0.8}{0.178}} & \secondres{\scalebox{0.8}{0.270}} &\scalebox{0.8}{0.219} &\scalebox{0.8}{0.298} & \scalebox{0.8}{0.205} & \scalebox{0.8}{0.290} & \scalebox{0.8}{0.244} & \scalebox{0.8}{0.334}  & \scalebox{0.8}{0.251} & \scalebox{0.8}{0.344} &\scalebox{0.8}{0.192} &\scalebox{0.8}{0.295} &\scalebox{0.8}{0.212} &\scalebox{0.8}{0.300} & \scalebox{0.8}{0.268} & \scalebox{0.8}{0.365} &\scalebox{0.8}{0.214} &\scalebox{0.8}{0.327} &{\scalebox{0.8}{0.193}} &{\scalebox{0.8}{0.296}} &\scalebox{0.8}{0.227} &\scalebox{0.8}{0.338} \\
    \midrule
    \scalebox{0.95}{\update{ETTh1}} &
    
    {{\scalebox{0.8}{0.443}}} 
    & \secondres{{\scalebox{0.8}{0.432}}} 

     & \secondres{{\scalebox{0.8}{0.428}}} & {\scalebox{0.8}{0.435}}
      & \boldres{{\scalebox{0.8}{0.426}}} & \boldres{{\scalebox{0.8}{0.431}}} 
       & {\scalebox{0.8}{0.500}} & {\scalebox{0.8}{0.474}} 
    & {\scalebox{0.8}{0.450}} & {\scalebox{0.8}{0.439}} 
    
    & {\scalebox{0.8}{0.454}} & {\scalebox{0.8}{0.447}} & 
    
    {\scalebox{0.8}{{0.446}}} & {\scalebox{0.8}{0.434}} & 
    
    \scalebox{0.8}{0.469} & 
    \scalebox{0.8}{0.454} &
    
    \scalebox{0.8}{{0.529}} & 
    \scalebox{0.8}{{0.522}} 
    
    & \scalebox{0.8}{{0.541}} & \scalebox{0.8}{{0.507}}
    
    & \scalebox{0.8}{{0.458}} & \scalebox{0.8}{{0.450}} 
    
    & \scalebox{0.8}{{0.456}} & \scalebox{0.8}{{0.452}} 
    
    & \scalebox{0.8}{{0.747}} & \scalebox{0.8}{{0.647}} 
    
    & {\scalebox{0.8}{{0.440}}} & \scalebox{0.8}{{0.460}} 
    
    & \scalebox{0.8}{{0.570}} & \scalebox{0.8}{{0.537}}
    
    & \scalebox{0.8}{{0.496}} & \scalebox{0.8}{{0.487}} \\
    \midrule
    \scalebox{0.95}{\update{ETTh2}} & 
    \boldres{{\scalebox{0.8}{0.363}}} & \boldres{{\scalebox{0.8}{0.392}}}& 
     \secondres{\scalebox{0.8}{0.372}} & {\scalebox{0.8}{0.401}}
     & 
     {\scalebox{0.8}{0.383}} & {\scalebox{0.8}{0.416}}
     & 
     {\scalebox{0.8}{0.398}} & {\scalebox{0.8}{0.416}}&
    {\scalebox{0.8}{0.385}} & {\scalebox{0.8}{0.411}}&
    
    {\scalebox{0.8}{0.383}} & {\scalebox{0.8}{0.407}} 
    
    & {\scalebox{0.8}{{0.374}}} & \secondres{\scalebox{0.8}{0.398}} & 
    
    \scalebox{0.8}{0.387} & 
    \scalebox{0.8}{0.407} & 
    
    \scalebox{0.8}{{0.942}} & \scalebox{0.8}{{0.684}} & 
    
    \scalebox{0.8}{{0.611}} & \scalebox{0.8}{{0.550}} 
    
    & \scalebox{0.8}{{0.414}} & \scalebox{0.8}{{0.427}} 
    
    & \scalebox{0.8}{{0.559}} & \scalebox{0.8}{{0.515}} 
    
    & \scalebox{0.8}{{0.954}} & \scalebox{0.8}{{0.723}} 
    
    & \scalebox{0.8}{{0.437}} & \scalebox{0.8}{{0.449}} 
    
    & \scalebox{0.8}{{0.526}} & \scalebox{0.8}{{0.516}} 
    
    & \scalebox{0.8}{{0.450}} & \scalebox{0.8}{{0.459}} \\
    
    \midrule
    \scalebox{0.95}{\update{ETTm1}} 
    & \boldres{{\scalebox{0.8}{0.378}}} & \boldres{{\scalebox{0.8}{0.394}}} 
    & {\scalebox{0.8}{0.388}} & {\scalebox{0.8}{0.405}} 
     & {\scalebox{0.8}{0.393}} & {\scalebox{0.8}{0.408}} 
      & {\scalebox{0.8}{0.409}} & {\scalebox{0.8}{0.407}} 
    & {\scalebox{0.8}{0.392}} & {\scalebox{0.8}{0.401}} 
    
    & {\scalebox{0.8}{0.407}} & {\scalebox{0.8}{0.410}} & 
    
    \scalebox{0.8}{{0.414}} & \scalebox{0.8}{0.407} & 
    
    \secondres{\scalebox{0.8}{0.387}} & \secondres{\scalebox{0.8}{0.400}} & 
    
    \scalebox{0.8}{{0.513}} & \scalebox{0.8}{{0.496}} 
    
    & \scalebox{0.8}{{0.419}} &
    \scalebox{0.8}{{0.419}} 
    
    & \scalebox{0.8}{{0.400}} & 
    \scalebox{0.8}{{0.406}} 
    
    & \scalebox{0.8}{{0.403}} 
    & \scalebox{0.8}{{0.407}} & 
    
    \scalebox{0.8}{{0.485}} 
    & \scalebox{0.8}{{0.481}} & 
    
    \scalebox{0.8}{{0.448}}   
    & \scalebox{0.8}{{0.452}} & 
    
    \scalebox{0.8}{{0.481}} 
     & \scalebox{0.8}{{0.456}} & 
     
     \scalebox{0.8}{{0.588}} & \scalebox{0.8}{{0.517}} \\

    \midrule 
    \scalebox{0.95}{\update{ETTm2}}
    & \secondres{{\scalebox{0.8}{0.276}}} & \boldres{{\scalebox{0.8}{0.322}}}& 
    \boldres{\scalebox{0.8}{0.275}} & \boldres{\scalebox{0.8}{0.322}}&
    {\scalebox{0.8}{0.283}} & {\scalebox{0.8}{0.335}}&
    {\scalebox{0.8}{0.288}} & {\scalebox{0.8}{0.332}}&
    {\scalebox{0.8}{0.285}} & {\scalebox{0.8}{0.331}}& 
    
    {\scalebox{0.8}{0.288}} & {\scalebox{0.8}{0.332}} & 
    
    \scalebox{0.8}{{0.286}} &\scalebox{0.8}{0.327} & 
    
    {\scalebox{0.8}{0.281}} & \secondres{\scalebox{0.8}{0.326}} & 
    
    \scalebox{0.8}{{0.757}} & \scalebox{0.8}{{0.610}}
    
    & \scalebox{0.8}{{0.358}} & \scalebox{0.8}{{0.404}} 
    
    & \scalebox{0.8}{{0.291}} & \scalebox{0.8}{{0.333}} 
    
    & \scalebox{0.8}{{0.350}} & \scalebox{0.8}{{0.401}} 
    
    & \scalebox{0.8}{{0.571}} & \scalebox{0.8}{{0.537}}
    
    & \scalebox{0.8}{{0.305}} & \scalebox{0.8}{{0.349}} 
    
    & \scalebox{0.8}{{0.306}} & \scalebox{0.8}{{0.347}} 
    
    & \scalebox{0.8}{{0.327}} & \scalebox{0.8}{{0.371}} \\
    \midrule 
    \scalebox{0.95}{\update{Exchange}} 
    &\secondres{\scalebox{0.8}{0.356}} & \boldres{\scalebox{0.8}{0.401}}
    &\scalebox{0.8}{0.378} & \scalebox{0.8}{0.411}
    &\scalebox{0.8}{0.382} & \scalebox{0.8}{0.412}
    &\scalebox{0.8}{0.395} & \scalebox{0.8}{0.421}
    &\scalebox{0.8}{0.368} & \scalebox{0.8}{0.406}
    & {\scalebox{0.8}{0.360}} & \secondres{\scalebox{0.8}{0.403}} &\scalebox{0.8}{0.378} &\scalebox{0.8}{0.417} & \scalebox{0.8}{0.367} & {\scalebox{0.8}{0.404}} & \scalebox{0.8}{0.940} & \scalebox{0.8}{0.707} & \scalebox{0.8}{0.370} & \scalebox{0.8}{0.413} &{\scalebox{0.8}{0.416}} &{\scalebox{0.8}{0.443}} & \boldres{\scalebox{0.8}{0.354}} &\scalebox{0.8}{0.414} & \scalebox{0.8}{0.750} & \scalebox{0.8}{0.626} &{\scalebox{0.8}{0.519}} &\scalebox{0.8}{0.429} &\scalebox{0.8}{0.461} &{\scalebox{0.8}{0.454}} &\scalebox{0.8}{0.613} &\scalebox{0.8}{0.539} \\

    \midrule 
    \scalebox{0.95}{Traffic} & \boldres{\scalebox{0.8}{0.422}} & \boldres{\scalebox{0.8}{0.276}}& 
     \scalebox{0.8}{0.622} & \scalebox{0.8}{0.309}& 
     \scalebox{0.8}{0.670} & \scalebox{0.8}{0.332}& 
     \scalebox{0.8}{0.481} & \scalebox{0.8}{0.321}& 
     
    \scalebox{0.8}{0.511} & \scalebox{0.8}{0.334}& 
    
    \secondres{\scalebox{0.8}{0.428}} & \secondres{\scalebox{0.8}{0.282}} &\scalebox{0.8}{0.626} &\scalebox{0.8}{0.378} & \scalebox{0.8}{0.481} & \scalebox{0.8}{0.304} & \scalebox{0.8}{0.550} & \scalebox{0.8}{0.304} & \scalebox{0.8}{0.760} & \scalebox{0.8}{0.473} &{\scalebox{0.8}{0.620}} &{\scalebox{0.8}{0.336}} &\scalebox{0.8}{0.625} &\scalebox{0.8}{0.383} & \scalebox{0.8}{0.804} & \scalebox{0.8}{0.509} &{\scalebox{0.8}{0.610}} &\scalebox{0.8}{0.376} &\scalebox{0.8}{0.624} &{\scalebox{0.8}{0.340}} &\scalebox{0.8}{0.628} &\scalebox{0.8}{0.379} \\
    
    \midrule
    \scalebox{0.95}{Weather} & \boldres{\scalebox{0.8}{0.249}} & \secondres{\scalebox{0.8}{0.278}} 
    &\secondres{\scalebox{0.9}{0.250}} & {\scalebox{0.8}{0.298}}
    &{{\scalebox{0.9}{0.256}}} & {\scalebox{0.8}{0.302}}
    &\boldres{\scalebox{0.9}{0.249}} & \boldres{\scalebox{0.8}{0.276}}&
    \scalebox{0.8}{0.267} & \scalebox{0.8}{0.287} 
    
    & {\scalebox{0.8}{0.258}} & \secondres{\scalebox{0.8}{0.278}} &\scalebox{0.8}{0.272} &\scalebox{0.8}{0.291} & \scalebox{0.8}{0.259} & \scalebox{0.8}{0.281} & \scalebox{0.8}{0.259} & \scalebox{0.8}{0.315}  & \scalebox{0.8}{0.271} & \scalebox{0.8}{0.320} &{\scalebox{0.8}{0.259}} &{\scalebox{0.8}{0.287}} &\scalebox{0.8}{0.265} &\scalebox{0.8}{0.317} & \scalebox{0.8}{0.292} & \scalebox{0.8}{0.363} &\scalebox{0.8}{0.309} &\scalebox{0.8}{0.360} &\scalebox{0.8}{0.288} &\scalebox{0.8}{0.314} &\scalebox{0.8}{0.338} &\scalebox{0.8}{0.382} \\
    \midrule
    \scalebox{0.95}{Solar-Energy}
      &\secondres{\scalebox{0.8}{0.229}} & \boldres{\scalebox{0.8}{0.253}}
    &\boldres{\scalebox{0.8}{0.225}} & {\scalebox{0.8}{0.266}}
     &\scalebox{0.8}{0.239} & \scalebox{0.8}{0.282}
      &\scalebox{0.8}{0.245} & \scalebox{0.8}{0.277}
    &\scalebox{0.8}{0.269} & \scalebox{0.8}{0.304}
    &{\scalebox{0.8}{0.233}} &\secondres{\scalebox{0.8}{0.262}} &\scalebox{0.8}{0.369} &\scalebox{0.8}{0.356} &{\scalebox{0.8}{0.270}} &{\scalebox{0.8}{0.307}} &\scalebox{0.8}{0.641} &\scalebox{0.8}{0.639} &\scalebox{0.8}{0.347} &\scalebox{0.8}{0.417} &\scalebox{0.8}{0.301} &\scalebox{0.8}{0.319} &\scalebox{0.8}{0.330} &\scalebox{0.8}{0.401} &\scalebox{0.8}{0.282} &\scalebox{0.8}{0.375} &\scalebox{0.8}{0.291} &\scalebox{0.8}{0.381} &\scalebox{0.8}{0.261} &\scalebox{0.8}{0.381} &\scalebox{0.8}{0.885} &\scalebox{0.8}{0.711} \\
    

    \bottomrule
  \end{tabular}
    \end{small}
  \end{threeparttable}
}
\end{table*}

\subsection{Ablation Study}
\begin{figure}[htbp]
\begin{center}
\includegraphics[width=0.95\columnwidth]{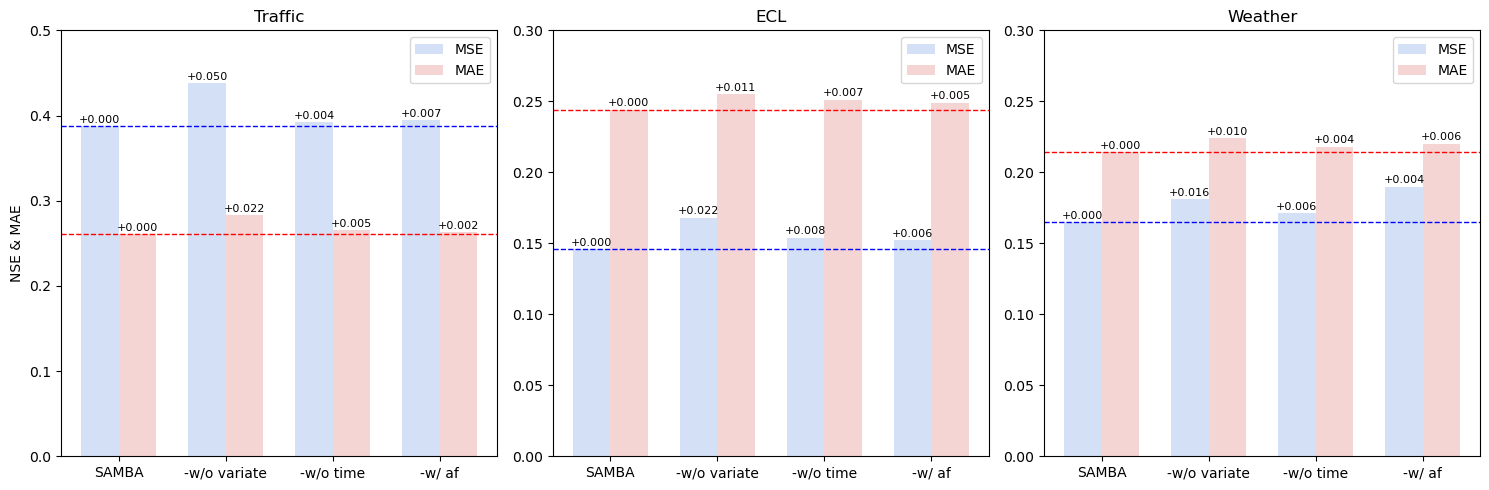}
\caption{\update{Ablation study of removing nonlinear activation function and disentangled encoding strategy in \model. 4 cases are include: 1) \model; 2) \model without time encoding; 3) \model without variate encoding; 4) Mamba with disentangled encoding.}}
\label{fig:ablation_resluts}
\end{center}
\end{figure}

To verify the rationale for removing the non-linear activation function and the disentangled encoding strategy, we choose our proposed \model as the SOTA benchmark and develop three variant models. Figure \ref{fig:ablation_resluts} shows that \model consistently outperforms other variants. Excluding either the nonlinear activation function or any branch of the disentangled encoding strategy results in a performance decline, demonstrating the effectiveness of our architecture and strategy. After incorporating the patch technique, the impact of the nonlinear activation function decreased. This occurs because the patch technique enriches the information available for each token, thereby enhancing the capture of semantic dependency. Additionally, our analysis shows that removing the cross-variate dependency encoding branch leads to a more significant performance degradation on the Traffic dataset compared to other datasets. This result is intuitive, as variates in traffic scenarios typically exhibit strong interdependencies \cite{traffic}.


\subsection{Evaluating the Universality of Disentangled Encoding Strategy}
We utilize Transformer and their variants, including Informer \cite{zhou2021informer} and PatchTST \cite{patchtst}, to validate the universality of our proposed disentangled encoding strategy. Table \ref{tab:forecasting_promotion} has demonstrated our strategy is effective not only for Mamba but also for Transformer-based models. Significant improvements indicate that these models have not properly modeled and introduced cross-variate dependency, whereas our strategy effectively mitigates these issues. Therefore, this disentangled encoding strategy is a model-agnostic method and could potentially be integrated with other models.

\subsection{Efficiency Analysis of SDE-Mamba}
\label{section:Efficiency}
First, the time complexity of each Mamba block is $O(D^2 T + D T \log T)$ (~\cite{fu2022hungry,gu2023mamba}), where $D$ is the dimension of token embeddings and $T$ is the length of tokens. As a \model block only removes the nonlinear activation before the selection mechanism, it possesses the same time complexity. Next, we consider the time complexity of the disentangled modeling.
(i) The linear projection in the tokenization process takes time $O(NJPD).$
(ii) In the CI learning, the token length is the variate number $N,$ thereby taking time $O(N(D^2 J + D J \log J)).$
(iii) In the CD learning, the dimension of token embeddings becomes the patch number $J,$ while the token length becomes the hidden dimension $D.$ Considering the bidirectional design, it takes time $O(2J(D^2 N + D N \log N)).$ 
(iv) The FFN-based aggregation operation and the final prediction head take time $O(NJD^2 + NJDS).$ In general, the time complexity of \model is $O(NJPD + N(D^2 J + D J \log J) + 2J(D^2 N + D N \log N) + (NJD^2 + NJDS)) = O(D^2 \cdot NJ + D \cdot NJ \log (NJ)).$ This validates that \model maintains the \emph{near linear} time complexity \wrt of the patch number and the variate number, a key merit of Mamba, while also inheriting the efficiency of the patching strategy.


In addition, we provide a detailed efficiency comparison between \model and the baseline models. As shown in Table~\ref{tab:cost}, \model achieves both a faster training speed and a smaller memory usage compared to many SOTA transformer-based models, such as PatchTST and Crossformer, which also employ attention mechanisms in temporal dimensions. Furthermore, \model exhibits a much smaller increase in memory consumption and training time as input lengths grow, underscoring its superior overall efficiency. 

\subsection{Short-Term Forecasting}
\begin{table}[htbp]
\label{tab:short_term}
\centering
\caption{Performance comparison of \model and ablation variants. 4 cases are include: 1) \model; 2) \model without time encoding; 3) \model without variate encoding; 4) Mamba with disentangled encoding.}
\resizebox{\linewidth}{!}{%
\begin{tabular}{l|ccccc}
\toprule
 & \textbf{SDE-Mamba} & \textbf{w/o variat} & \textbf{w/o time} & \textbf{w/ af} & \textbf{iTransformer} \\
\midrule
\textbf{SMAPE} & \textbf{12.129} & 12.492 & 12.279 & 12.323 & 12.706 \\
\textbf{MASE}  & \textbf{14.383} & 14.876 & 14.632 & 14.590 & 15.269 \\
\textbf{MAPE}  & \textbf{1.652}  & 1.697  & 1.661  & 1.693  & 1.737 \\
\textbf{OWA}   & \textbf{0.878}  & 0.904  & 0.887  & 0.897  & 0.923 \\
\bottomrule
\end{tabular}%
}
\label{tab:sde-maml-ablation}
\end{table}
In addition to long-term forecasting (LTSF), we also evaluate the proposed SDE-Mamba framework on a short-term forecasting task using the M4 dataset \cite{makridakis2018m4}, which comprises 100,000 univariate time series from marketing domains sampled at yearly, quarterly, and monthly frequencies. As shown in Table~\ref{tab:sde-maml-ablation}, the full SDE-Mamba model consistently achieves the best performance across all four metrics, demonstrating its strong generalization ability beyond the LTSF setting. We further conduct ablation studies to assess the role of each component. All variants show performance drops in most metrics, highlighting the necessity of both time and variate encoding for modeling temporal and feature-specific patterns.



\begin{table}[htbp]
  \centering
  \vskip 0.05in
  \begin{minipage}[t]{0.45\textwidth} 
    \centering
    \large
    \resizebox{\textwidth}{!}{ 
      \begin{threeparttable}
        \caption{Performance gains of our disentangled encoding strategy on MAE and MSE.} \label{tab:forecasting_promotion}
        \vskip 0.05in
        \begin{small}
          \renewcommand{\multirowsetup}{\centering}
          \setlength{\tabcolsep}{2.2pt} 
          \begin{tabular}{c|c|cc|cc|cc}
            \toprule
            \multicolumn{2}{c|}{\multirow{2}{*}{{Models}}} & 
            \multicolumn{2}{c}{Transformer} &
            \multicolumn{2}{c}{PatchTST} &
            \multicolumn{2}{c}{Informer} \\
            \cmidrule(lr){3-4} \cmidrule(lr){5-6} \cmidrule(lr){7-8} 
            \multicolumn{2}{c|}{Metric} & MSE & MAE & MSE & MAE & MSE & MAE \\
            \toprule
            \multirow{3}{*}{ETTh1} & Original & 0.997 & 0.797 & 0.469 & 0.454 & 1.060 & 0.791 \\
            & \textbf{+disentangled} & \textbf{0.481} & \textbf{0.477} & \textbf{0.450} & \textbf{0.439} & \textbf{0.489} & \textbf{0.485} \\
            \cmidrule(lr){2-8}
            & Promotion & 51.8\% & 42.2\% & 4.05\% & 3.31\% & 53.9\% & 38.7\% \\
            \midrule
            \multirow{3}{*}{ETTm1} & Original & 0.773 & 0.656 & 0.387 & 0.400 & 0.870 & 0.696 \\
            & \textbf{+disentangled} & \textbf{0.454} & \textbf{0.461} & \textbf{0.383} & \textbf{0.396} & \textbf{0.470} & \textbf{0.459} \\
            \cmidrule(lr){2-8}
            & Promotion & 41.3\% & 29.7\% & 1.03\% & 1.00\% & 46.0\% & 34.1\% \\
            \midrule
            \multirow{3}{*}{Weather} & Original & 0.657 & 0.572 & 0.259 & 0.281 & 0.634 & 0.548 \\
            & \textbf{+disentangled} & \textbf{0.258} & \textbf{0.279} & \textbf{0.248} & \textbf{0.278} & \textbf{0.271} & \textbf{0.330} \\
            \cmidrule(lr){2-8}
            & Promotion & 60.7\% & 51.2\% & 4.25\% & 1.08\% & 57.3\% & 39.8\% \\
            \bottomrule
          \end{tabular}
        \end{small}
      \end{threeparttable}
    }
  \end{minipage}%
  \hspace{2\linewidth} 
  \begin{minipage}[t]{0.49\textwidth} 
    \centering
    \large
    \resizebox{\textwidth}{!}{ 
      \begin{threeparttable}
        \caption{Efficiency Analysis: The GPU memory (MiB) and speed (running time, s/iter) of each model on Traffic dataset. Mem means memory footprint.}\label{tab:cost}
        \vskip 0.05in
        \begin{small}
          \renewcommand{\multirowsetup}{\centering}
          \setlength{\tabcolsep}{6pt} 
          \begin{tabular}{ll|cccccc}
            \toprule
            \multicolumn{2}{l|}{Input Length} &\multicolumn{2}{c|}{96} &\multicolumn{2}{c|}{336}  &\multicolumn{2}{c}{720}\\
            \cmidrule(lr){3-4} \cmidrule(lr){5-6} \cmidrule(lr){7-8}
            \multicolumn{2}{l|}{Models}  & Mem & Speed  & Mem & Speed & Mem & Speed \\
            \toprule
            \multicolumn{2}{l|}{SDE-Mamba}     & 2235  & 0.0403  & 2275  & 0.0711  & 2311  & 0.1232 \\
      \multicolumn{2}{l|}{PatchTST}      & 3065  & 0.0658  & 12299 & 0.2382  & 25023 & 0.4845 \\
      \multicolumn{2}{l|}{iTransformer}  & 3367  & 0.0456  & 3389  & 0.0465  & 3411  & 0.0482 \\
      \multicolumn{2}{l|}{DLinear}       & 579   & 0.0057  & 619   & 0.0082  & 681   & 0.0139 \\
      \multicolumn{2}{l|}{TimesNet}      & 6891  & 0.2492  & 7493  & 0.4059  & 9807  & 0.6289 \\
      \multicolumn{2}{l|}{Crossformer}   & 21899 & 0.7941  & 40895 & 1.426  & 69771 & 2.424 \\
      \multicolumn{2}{l|}{FEDFormer}     & 1951  & 0.1356  & 1957  & 0.1369  & 2339  & 0.1643 \\
      \multicolumn{2}{l|}{Autoformer}    & 1489  & 0.0309  & 1817  & 0.0362  & 2799  & 0.0457 \\
            \bottomrule
          \end{tabular}
        \end{small}
      \end{threeparttable}
    }
  \end{minipage} 
\end{table}
\subsection{Increasing Lookback Length}
\begin{figure}[htbp]
\begin{center}
\includegraphics[width=1.0\columnwidth]{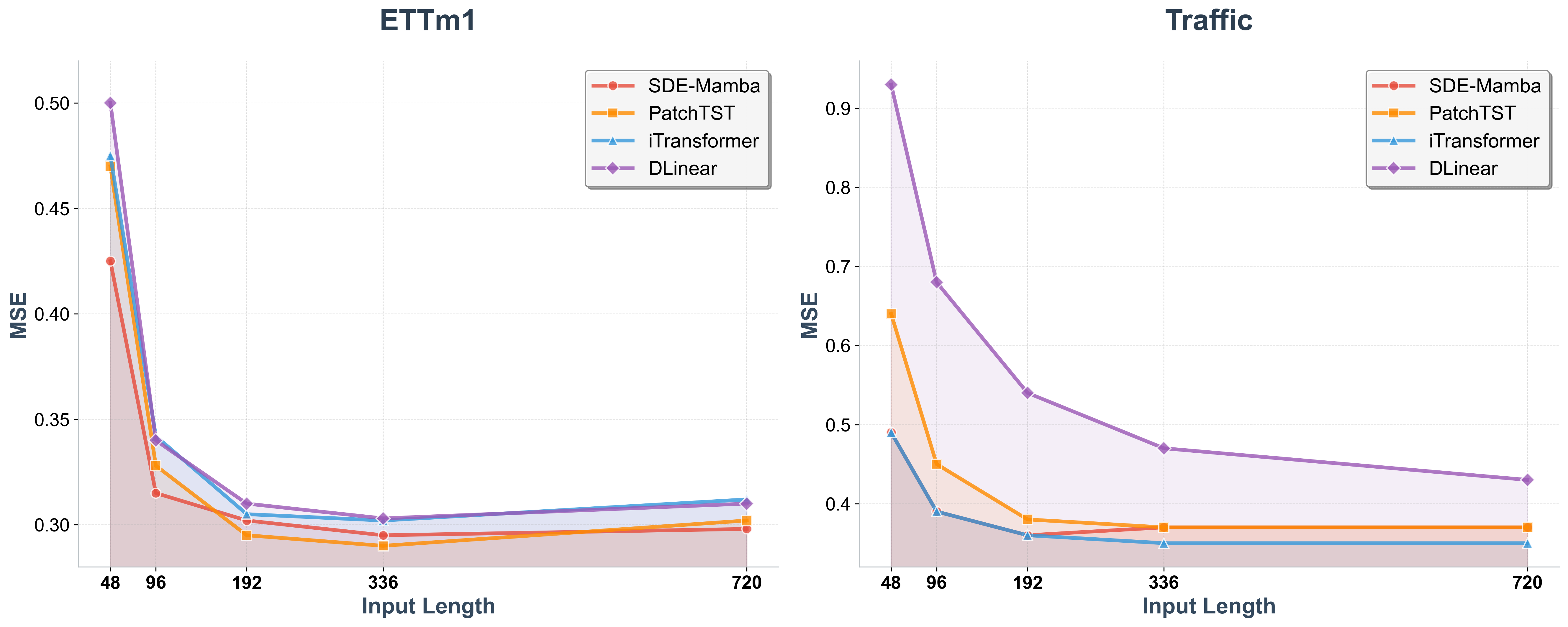}


\caption{
Forecasting performance with the lookback length $T \in \{48, 96, 192, 336, 720\}$ and fixed prediction length $S = 96$. 
}
\label{fig:increase}
\end{center}
\end{figure}

It is intuitive to expect that using a longer lookback window would enhance forecasting performance. However, previous research has shown that most transformer-based models cannot leverage the increased receptive field to improve predictive accuracy effectively (\cite{patchtst}). As a result, SOTA transformer-based models, such as iTransformer and PatchTST, have improved their design to more effectively utilize a longer lookback window. In evaluating \model, we explore two key questions: (1) Can \model effectively take advantage of a longer lookback window to achieve more accurate forecasting? (2) If so, how does its performance compare to other models that can also utilize a longer lookback window? For comparison, we select four baseline models—DLinear, PatchTST, and iTransformer. These models have been proven to utilize extended a lookback window effectively (\cite{liu2023itransformer, patchtst}). As shown in Figure \ref{fig:increase}, we observe a steady decrease in MSE as the lookback window increases, confirming \model’s capacity to learn from longer temporal contexts. Moreover, while both \model and the baseline models can effectively take advantage of a longer look-back window, \model consistently demonstrates superior overall performance.
\vspace{-10pt}
\section{Conclusions and Future Work}
In this paper, we identify and rigorously define three crucial dependencies in time series: order dependency, semantic dependency, and cross-variate dependency. Based on these dependencies, we examine the three limitations of existing time series forecasting models: (1) The inability to capture order and semantic dependencies simultaneously. (2) The presence of overfitting issues. (3) The improper introduction of cross-variate dependency. We further empirically verify the superiority of state space model as an alternative to linear and Transformer models, despite its redundant nonlinear activation functions. Building on these insights, we propose SDE, a novel framework designed to enhance the capability of SSMs
for LTSF. Experimentally, \model achieves SOTA performance, validating the rationality and efficacy of our method design grounded in rich empirical findings. Our efforts further demonstrate the great potential of state space models in time series modeling. In the future, we aim to delve deeper into diverse time series tasks, pushing the boundaries of analysis with state space model. 

\begin{acks}
This work was supported by the National Key R\&D Program of China (Grant No.2023YFF0725004), National Natural Science Foundation of China (Grant No.92370204), the Guangzhou Basic and Applied Basic Research Program under Grant No. 2024A04J3279, Education Bureau of Guangzhou Municipality, and CCF-DiDi GAIA Collaborative Research Funds.
\end{acks}


\newpage
\bibliographystyle{ACM-Reference-Format}
\bibliography{sample-base}


\appendix

\section{Theoretical Discussions}
\label{app:thm}

\subsection{Proof for Theorem 1}
Without loss of generality, we only consider the setting with two variates, \ie $N=2,$ and treat the 1st variate as the target variable. For simplicity, we introduce the notation $\Psi \left( \left(\mathbf{A}, \mathbf{B}\right)^T \right) := \Psi(\mathbf{A}, \mathbf{B}).$ Then, we can express the outputs of $\mathcal{F}_d(\cdot)$ and $\mathcal{F}_{ttv}(\cdot)$ \wrt the target variate as follows:
\begin{align}
    \mathcal{F}_d(\mathbf{E})^1 &= \text{FFN} \left( \Phi(\mathbf{E}^1) || \Psi \left( \mathbf{E}^1, \mathbf{E}^2 \right)^1 \right), \\
    \mathcal{F}_{ttv}(\mathbf{E})^1 &= 
    \text{FFN}\left( \Psi \left( \Phi(\mathbf{E}^1), \Phi(\mathbf{E}^2) \right)^1 \right).
\end{align}

Next, we prove the informativeness of output representation from $\mathcal{F}_d$ compared to that of $\mathcal{F}_{ttv}.$
Based on the information loss property of data processing, we have
\begin{align}
    \begin{split}
        H(\mathbf{c}^1|\mathcal{F}_{ttv}(\mathbf{E})^1) &= H\left(\mathbf{c}^1|\text{FFN}\left( \Psi \left( \Phi(\mathbf{E}^1), \Phi(\mathbf{E}^2) \right)^1 \right) \right) \\
        & \ge H\left(\mathbf{c}^1| \text{FFN}\left( \Psi \left( \Phi(\mathbf{E}^1), \Phi(\mathbf{E}^2) \right)^1 \right) \right)\\
        &= H\left( \mathbf{c}^1| \text{FFN}\left( \Psi \left( \Phi(\mathbf{E}^1), \phi(\mathcal{O}(\mathbf{E}^2), \mathcal{S}(\mathbf{E}^2)) + \mathcal{Z}(\mathbf{E}^2) \right)^1 \right) \right) \\
        & \ge H\left( \mathbf{c}^1| \text{FFN}\left( \Psi \left( \Phi(\mathbf{E}^1), \mathcal{O}(\mathbf{E}^2) \right)^1\right), \mathcal{S}(\mathbf{E}^2), \mathcal{Z}(\mathbf{E}^2) \right).
    \end{split}
\end{align} 
Given the assumption (1) on the noisy components, 
we have
\begin{align}
    H(\mathbf{c}^1|\mathcal{F}_{ttv}(\mathbf{E})^1) \ge H\left(\mathbf{c}^1|\text{FFN}\left( \Psi \left( \Phi(\mathbf{E}^1), \mathcal{O}(\mathbf{E}^2) \right)^1\right), \mathcal{S}(\mathbf{E}^2) \right).
\end{align}

Furthermore, by combining assumption (2) on the semantic dependencies, we have
\begin{align}
    \begin{split}
        H(\mathbf{c}^1|\mathcal{F}_{ttv}(\mathbf{E})^1) &\ge H\left(\mathbf{c}^1|\text{FFN} \left( \Psi \left( \Phi(\mathbf{E}^1), \mathcal{O}(\mathbf{E}^2) \right)^1 \right), \mathcal{S}(\mathbf{E}^2) \right) \\
        &= H\left(\mathbf{c}^1|\text{FFN} \left( \Psi \left( \Phi(\mathbf{E}^1), \mathcal{O}(\mathbf{E}^2) \right)^1 \right) \right) \\
        &- I\left(\mathbf{c}^1; \mathcal{S}(\mathbf{E}^2) | \text{FFN} \left( \Psi \left( \Phi(\mathbf{E}^1), \mathcal{O}(\mathbf{E}^2) \right)^1 \right) \right) \\
        &= H\left(\mathbf{c}^1|\text{FFN} \left( \Psi \left( \Phi(\mathbf{E}^1), \mathcal{O}(\mathbf{E}^2) \right)^1 \right) \right) \\
        &- I(\mathbf{c}^1; \mathcal{S}(\mathbf{E}^2) | \Phi(\mathbf{E}^1), \mathcal{O}(\mathbf{E}^2)) \\
        &\ge H\left(\mathbf{c}^1|\text{FFN} \left( \Psi \left( \Phi(\mathbf{E}^1), \mathcal{O}(\mathbf{E}^2) \right)^1 \right) \right) - \epsilon \\
        &\ge H\left(\mathbf{c}^1|\text{FFN} \left( \Phi(\mathbf{E}^1) || \Psi \left( \Phi(\mathbf{E}^1), \mathcal{O}(\mathbf{E}^2) \right)^1 \right) \right) - \epsilon \\
        &\ge H\left(\mathbf{c}^1|\text{FFN} \left( \Phi(\mathbf{E}^1) || \mathcal{O}(\mathbf{E}^2) \right) \right) - \epsilon.
    \end{split}
\end{align}

For $\mathcal{F}_d,$ though the cross-variate dependency encoding $\Psi(\cdot)$ does not explicitly model the order dependencies in $\mathbf{E}^2,$ it aggregates the patch embedding of the 2nd variate to the target variate at each time step. This enables the $\text{FFN}(\cdot)$ layer, which transforms the input with a linear mapping, to further capture the order dependencies. Moreover, according to the empirical findings in \tabref{tab:order_dependency_1} as well as previous works~\cite{DLinear,RLinear}, we assume that linear mapping is not worse than other complex cross-time encoders such as Transformers and Mamba in terms of order dependency modeling. Therefore, combining inequality (9), we have 
\begin{align}
    H(Y_1|F_d(\mathbf{E})^1) &= H(Y_1|\text{FFN} \left( \Phi(\mathbf{E}^1) || \Psi \left( \mathbf{E}^1, \mathbf{E}^2 \right)^1 \right)) \\
    &= H\left(Y_1|\text{FFN} \left( \Phi(\mathbf{E}^1)\right) || \text{FFN} \left(\Psi \left( \mathbf{E}^1, \mathbf{E}^2 \right)^1 \right) \right) \\
    &\le H(Y_1|\text{FFN} \left( \Phi(\mathbf{E}^1)\right) || \mathcal{O}(\mathbf{E}^2) ) \\
    &\le H(Y_1|\text{FFN} \left( \Phi(\mathbf{E}^1) || \mathcal{O}(\mathbf{E}^2) \right)) \\
    &\le H(Y_1|F_{ttv}(\mathbf{E})^1) + \epsilon.
\end{align}

\subsection{Rationality of the Assumption (2) in Theorem 1} \label{app:thm_assumption}
We interpret the rationality of the assumption from two perspectives: (1) When the lookback window is short, like the often used 1-hour window in traffic prediction task~\cite{wu2021autoformer}, there exist low-density semantic dependencies within each variate's observations, while the order dependencies from other variates could lead to a direct impact on the future of the target variate due to continuity in physical laws, \eg the increasing traffic volume spreads to neighboring road segments; (2) Even when the lookback window becomes longer, the widely adopted patching strategy compresses the most useful local semantics into a $D$-dim feature vector, which is attached to each time step of embedding $\mathbf{E}$. Consequently, the semantic dependencies extracted along the temporal dimension of $\mathbf{E}$ usually become less informative.
\section{Hyperparameter Sensitivity}
\begin{figure}[htbp]
\begin{center}
\includegraphics[width=1\columnwidth]{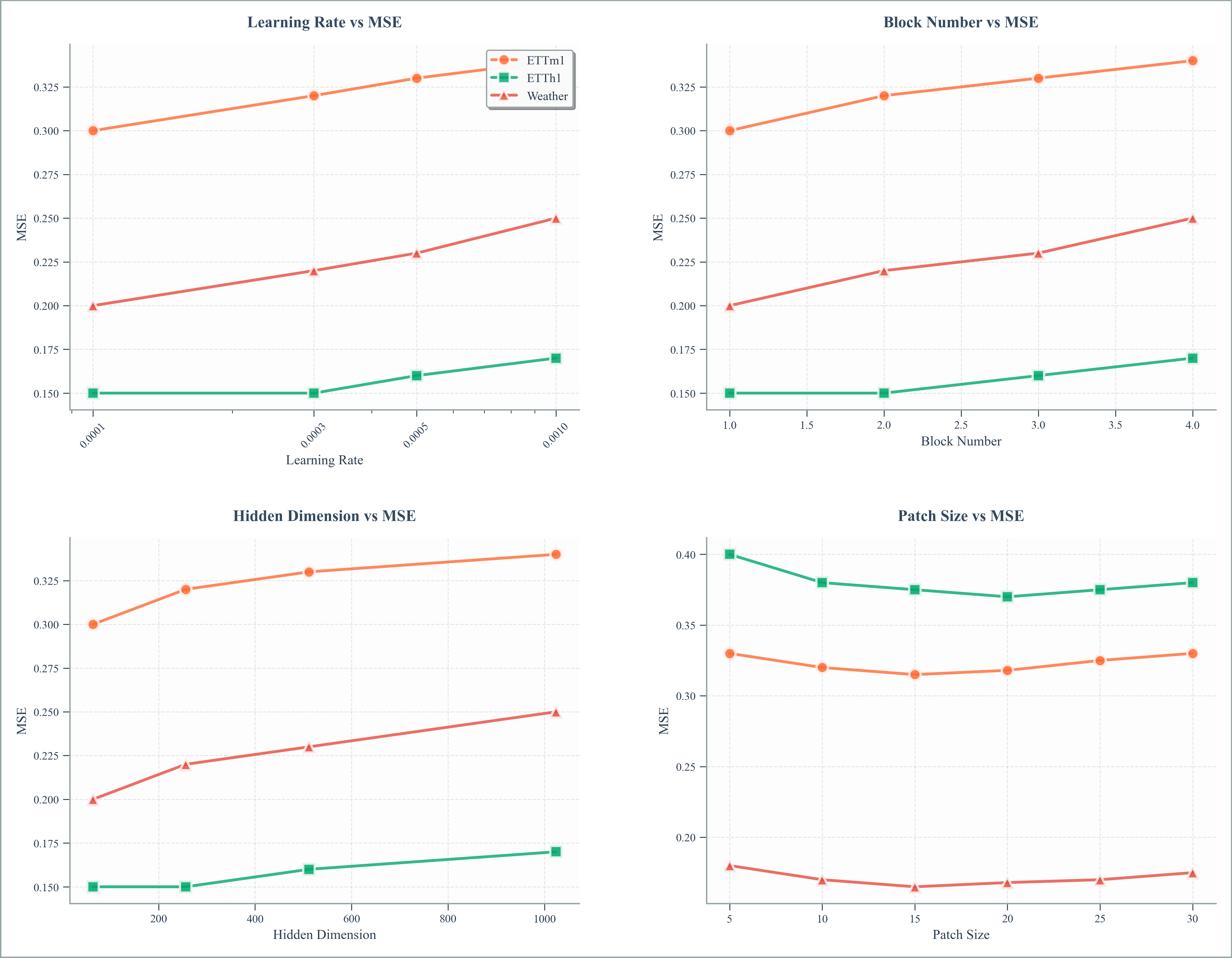}
\vspace{-5pt}
\caption{\update{ Hyperparameter sensitivity with respect to the learning rate, layer, dimension, and the patch size. The lookback window length $T=96$ and the forecast window length $S=96$
}}
\label{fig:hyp}
\end{center}
\vspace{-10pt}
\end{figure}
Figure \ref{fig:hyp} presents the Mean Squared Error (MSE) performance of three time series datasets (ETTm1, ETTh1, and Weather) across four key hyperparameters: learning rate, block number, hidden dimension, and patch size.  The results demonstrate that ETTm1 and Weather datasets show performance degradation with increasing learning rates and block numbers, while ETTh1 maintains relatively stable performance.  Hidden dimension scaling reveals consistent performance decline across all datasets, suggesting overfitting with larger dimensions.  Notably, patch size optimization exhibits distinct behaviors, with ETTh1 displaying a U-shaped curve optimal around size 20, while Weather achieves consistently superior performance with minimal sensitivity to patch variations.  Overall, our proposed method exhibits strong robustness across different hyperparameter configurations and maintains consistent performance even under suboptimal parameter settings.
\label{appendix:full_forecasting_results}
\begin{table*}[h]
  \caption{{Full results of the long-term forecasting task}. We compare extensive competitive models under different prediction lengths following the setting of TimesNet~\citeyearpar{wu2022timesnet}. The input sequence length is 96 for all baselines. \emph{Avg} means the average results from all four prediction lengths. }
  \label{tab:full_forecasting_results}
  \vskip -0.0in
  \vspace{3pt}
  \renewcommand{\arraystretch}{0.85} 
  \centering
  \resizebox{2\columnwidth}{!}{
  \begin{threeparttable}
  \begin{small}
  \renewcommand{\multirowsetup}{\centering}
  \setlength{\tabcolsep}{1pt}
  \begin{tabular}{c|c|cc|cc|cc|cc|cc|cc|cc|cc|cc|cc|cc|cc|cc|cc|cc|cc}
    \toprule
    \multicolumn{2}{c}{\multirow{2}{*}{Models}} & 
    \multicolumn{2}{c}{\rotatebox{0}{\scalebox{0.8}{\textbf{SDE-Mamba}}}} &
     \multicolumn{2}{c}{\rotatebox{0}{\scalebox{0.8}{\textbf{SDE-SegRNN}}}} &
      \multicolumn{2}{c}{\rotatebox{0}{\scalebox{0.8}{SegRNN}}} &
    \multicolumn{2}{c}{\rotatebox{0}{\scalebox{0.8}{CARD}}} &
    \multicolumn{2}{c}{\rotatebox{0}{\scalebox{0.8}{FTP}}} &
    \multicolumn{2}{c}{\rotatebox{0}{\scalebox{0.8}{iTransformer}}} &
    
    \multicolumn{2}{c}{\rotatebox{0}{\scalebox{0.8}{{RLinear}}}} &
    \multicolumn{2}{c}{\rotatebox{0}{\scalebox{0.8}{PatchTST}}} &
    \multicolumn{2}{c}{\rotatebox{0}{\scalebox{0.8}{Crossformer}}}  &
    \multicolumn{2}{c}{\rotatebox{0}{\scalebox{0.8}{TiDE}}} &
    \multicolumn{2}{c}{\rotatebox{0}{\scalebox{0.8}{{TimesNet}}}} &
    \multicolumn{2}{c}{\rotatebox{0}{\scalebox{0.8}{DLinear}}}&
    \multicolumn{2}{c}{\rotatebox{0}{\scalebox{0.8}{SCINet}}} &
    \multicolumn{2}{c}{\rotatebox{0}{\scalebox{0.8}{FEDformer}}} &
    \multicolumn{2}{c}{\rotatebox{0}{\scalebox{0.8}{Stationary}}} &
    \multicolumn{2}{c}{\rotatebox{0}{\scalebox{0.8}{Autoformer}}} \\
    \multicolumn{2}{c}{} &
     \multicolumn{2}{c}{\scalebox{0.8}{\textbf{Ours}}} &
       
       \multicolumn{2}{c}{\scalebox{0.8}{\textbf{Ours}}} &
       \multicolumn{2}{c}{\scalebox{0.8}{\citeyearpar{lin2023segrnn}}} &
      \multicolumn{2}{c}{\scalebox{0.8}{\citeyearpar{wang2024card}}} &
    \multicolumn{2}{c}{\scalebox{0.8}{\citeyearpar{GPT4TS}}} &
    \multicolumn{2}{c}{\scalebox{0.8}{\citeyearpar{liu2023itransformer}}} & 
    
    \multicolumn{2}{c}{\scalebox{0.8}{\citeyearpar{RLinear}}} & 
    \multicolumn{2}{c}{\scalebox{0.8}{\citeyearpar{patchtst}}} & 
    \multicolumn{2}{c}{\scalebox{0.8}{\citeyearpar{crossformer}}}  & 
    \multicolumn{2}{c}{\scalebox{0.8}{\citeyearpar{tide}}} & 
    \multicolumn{2}{c}{\scalebox{0.8}{\citeyearpar{wu2022timesnet}}} & 
    \multicolumn{2}{c}{\scalebox{0.8}{\citeyearpar{DLinear}}}& 
    \multicolumn{2}{c}{\scalebox{0.8}{\citeyearpar{SCINet}}} &
    \multicolumn{2}{c}{\scalebox{0.8}{\citeyearpar{fedformer}}} &
    \multicolumn{2}{c}{\scalebox{0.8}{\citeyearpar{Non_stationary}}} &
    \multicolumn{2}{c}{\scalebox{0.8}{\citeyearpar{wu2021autoformer}}} \\
    \cmidrule(lr){3-4} \cmidrule(lr){5-6}\cmidrule(lr){7-8} \cmidrule(lr){9-10}\cmidrule(lr){11-12}\cmidrule(lr){13-14} \cmidrule(lr){15-16} \cmidrule(lr){17-18} \cmidrule(lr){19-20} \cmidrule(lr){21-22} \cmidrule(lr){23-24} \cmidrule(lr){25-26} \cmidrule(lr){27-28} \cmidrule(lr){29-30} \cmidrule(lr){31-32} \cmidrule(lr){33-34}
    \multicolumn{2}{c}{Metric}  & \scalebox{0.78}{MSE} & \scalebox{0.78}{MAE}  & \scalebox{0.78}{MSE} & \scalebox{0.78}{MAE}  & \scalebox{0.78}{MSE} & \scalebox{0.78}{MAE}  & \scalebox{0.78}{MSE} & \scalebox{0.78}{MAE}  & \scalebox{0.78}{MSE} & \scalebox{0.78}{MAE} & \scalebox{0.78}{MSE} & \scalebox{0.78}{MAE}  &\scalebox{0.78}{MSE} & \scalebox{0.78}{MAE}  & \scalebox{0.78}{MSE} & \scalebox{0.78}{MAE} & \scalebox{0.78}{MSE} & \scalebox{0.78}{MAE} & \scalebox{0.78}{MSE} & \scalebox{0.78}{MAE} & \scalebox{0.78}{MSE} & \scalebox{0.78}{MAE} & \scalebox{0.78}{MSE} & \scalebox{0.78}{MAE} & \scalebox{0.78}{MSE} & \scalebox{0.78}{MAE} & \scalebox{0.78}{MSE} & \scalebox{0.78}{MAE}& \scalebox{0.78}{MSE} & \scalebox{0.78}{MAE} &\scalebox{0.78}{MSE} & \scalebox{0.78}{MAE}\\
    \toprule
    
    \multirow{4}{*}{{\rotatebox{90}{\scalebox{0.95}{ETTm1}}}}
    &  \scalebox{0.78}{96}  & \boldres{{\scalebox{0.78}{0.315}}} & \boldres{{\scalebox{0.78}{0.357}}} 
    &\scalebox{0.78}{0.330} &\scalebox{0.78}{0.371}
    &\scalebox{0.78}{0.336} &\scalebox{0.78}{0.375}
    &\scalebox{0.78}{0.351} &\scalebox{0.78}{0.379}&\scalebox{0.78}{0.335} & \scalebox{0.78}{0.367}
    & \scalebox{0.78}{0.334} & \scalebox{0.78}{0.368} & \scalebox{0.78}{0.355} & \scalebox{0.78}{0.376} &  \secondres{{\scalebox{0.78}{0.329}}} &  \secondres{{\scalebox{0.78}{0.367}}} & \scalebox{0.78}{0.404} & \scalebox{0.78}{0.426} & \scalebox{0.78}{0.364} & \scalebox{0.78}{0.387} &{\scalebox{0.78}{0.338}} &{\scalebox{0.78}{0.375}} &{\scalebox{0.78}{0.345}} &{\scalebox{0.78}{0.372}} & \scalebox{0.78}{0.418} & \scalebox{0.78}{0.438} &\scalebox{0.78}{0.379} &\scalebox{0.78}{0.419} &\scalebox{0.78}{0.386} &\scalebox{0.78}{0.398} &\scalebox{0.78}{0.505} &\scalebox{0.78}{0.475} \\ 
    
    & \scalebox{0.78}{192} & \boldres{{\scalebox{0.78}{0.360}}} & \boldres{{\scalebox{0.78}{0.383}}} 
    &\scalebox{0.78}{0.371} &\scalebox{0.78}{0.394}
    &\scalebox{0.78}{0.376} &\scalebox{0.78}{0.396}
    &\scalebox{0.78}{0.386} &\scalebox{0.78}{0.393}& \secondres{\scalebox{0.78}{0.366}} & \secondres{\scalebox{0.78}{0.385}}& \scalebox{0.78}{0.377} & \scalebox{0.78}{0.391} & \scalebox{0.78}{0.391} & \scalebox{0.78}{0.392} &  {\scalebox{0.78}{0.367}} &  \secondres{{\scalebox{0.78}{0.385}}} & \scalebox{0.78}{0.450} & \scalebox{0.78}{0.451} &\scalebox{0.78}{0.398} & \scalebox{0.78}{0.404} & {\scalebox{0.78}{0.374}} & {\scalebox{0.78}{0.387}}  &{\scalebox{0.78}{0.380}} &{\scalebox{0.78}{0.389}} & \scalebox{0.78}{0.439} & \scalebox{0.78}{0.450}  &\scalebox{0.78}{0.426} &\scalebox{0.78}{0.441} &\scalebox{0.78}{0.459} &\scalebox{0.78}{0.444} &\scalebox{0.78}{0.553} &\scalebox{0.78}{0.496} \\ 
    
    & \scalebox{0.78}{336} & \boldres{\scalebox{0.78}{0.389}} & \boldres{\scalebox{0.78}{0.405}}
    &\scalebox{0.78}{0.400} &\scalebox{0.78}{0.413}
    &\scalebox{0.78}{0.402} &\scalebox{0.78}{0.415}
    &\scalebox{0.78}{0.419} &\scalebox{0.78}{0.412}& \secondres{\scalebox{0.78}{0.398}} & \secondres{\scalebox{0.78}{0.407}} & \scalebox{0.78}{0.426} & \scalebox{0.78}{0.420} & \scalebox{0.78}{0.424} & \scalebox{0.78}{0.415} &  {\scalebox{0.78}{0.399}} &  {\scalebox{0.78}{0.410}} & \scalebox{0.78}{0.532}  &\scalebox{0.78}{0.515} & \scalebox{0.78}{0.428} & \scalebox{0.78}{0.425} & {\scalebox{0.78}{0.410}} & {\scalebox{0.78}{0.411}}  &{\scalebox{0.78}{0.413}} &{\scalebox{0.78}{0.413}} & \scalebox{0.78}{0.490} & \scalebox{0.78}{0.485}  &\scalebox{0.78}{0.445} &\scalebox{0.78}{0.459} &\scalebox{0.78}{0.495} &\scalebox{0.78}{0.464} &\scalebox{0.78}{0.621} &\scalebox{0.78}{0.537} \\ 
    
    & \scalebox{0.78}{720} & \boldres{\scalebox{0.78}{0.448}} & \secondres{\scalebox{0.78}{0.440}}
     &\secondres{\scalebox{0.78}{0.453}} &\scalebox{0.78}{0.443}
      &\scalebox{0.78}{0.458} &\scalebox{0.78}{0.445}
    &\scalebox{0.78}{0.479} &\scalebox{0.78}{0.445}& \scalebox{0.78}{0.461} & \scalebox{0.78}{0.443}& \scalebox{0.78}{0.491} & \scalebox{0.78}{0.459} & \scalebox{0.78}{0.487} & \scalebox{0.78}{0.450} &  {{\scalebox{0.78}{0.454}}} &  \boldres{{\scalebox{0.78}{0.439}}} & \scalebox{0.78}{0.666} & \scalebox{0.78}{0.589} & \scalebox{0.78}{0.487} & \scalebox{0.78}{0.461} &{\scalebox{0.78}{0.478}} & {\scalebox{0.78}{0.450}} & {\scalebox{0.78}{0.474}} &{\scalebox{0.78}{0.453}} & \scalebox{0.78}{0.595} & \scalebox{0.78}{0.550}  &\scalebox{0.78}{0.543} &\scalebox{0.78}{0.490} &\scalebox{0.78}{0.585} &\scalebox{0.78}{0.516} &\scalebox{0.78}{0.671} &\scalebox{0.78}{0.561} \\ 
    \cmidrule(lr){2-34}
    & \scalebox{0.78}{Avg} & \boldres{\scalebox{0.78}{0.378}} & \boldres{\scalebox{0.78}{0.394}}
    &\scalebox{0.78}{0.388} & \scalebox{0.78}{0.405} 
    &\scalebox{0.78}{0.393} & \scalebox{0.78}{0.408} 
    &\scalebox{0.78}{0.409} & \scalebox{0.78}{0.407} &\scalebox{0.78}{0.392} & \scalebox{0.78}{0.401} &\scalebox{0.78}{0.407} & \scalebox{0.78}{0.410} & \scalebox{0.78}{0.414} & \scalebox{0.78}{0.407} & \secondres{\scalebox{0.78}{0.387}} & \secondres{\scalebox{0.78}{0.400}} & \scalebox{0.78}{0.513} & \scalebox{0.78}{0.496} & \scalebox{0.78}{0.419} & \scalebox{0.78}{0.419} &\scalebox{0.78}{0.400} &{\scalebox{0.78}{0.406}}  &{\scalebox{0.78}{0.403}} &{\scalebox{0.78}{0.407}} & \scalebox{0.78}{0.485} & \scalebox{0.78}{0.481}  &\scalebox{0.78}{0.448} &\scalebox{0.78}{0.452} &\scalebox{0.78}{0.481} &\scalebox{0.78}{0.456} &\scalebox{0.78}{0.588} &\scalebox{0.78}{0.517} \\ 
    \midrule
    
    \multirow{4}{*}{{\rotatebox{90}{\scalebox{0.95}{ETTm2}}}}
    &  \scalebox{0.78}{96} &\secondres{{\scalebox{0.78}{0.172}}} & \secondres{{\scalebox{0.78}{0.259}}}
    &\boldres{\scalebox{0.78}{0.171}} &\boldres{\scalebox{0.78}{0.256}}&
    \scalebox{0.78}{0.176} &{\scalebox{0.78}{0.275}}&
    \scalebox{0.78}{0.185} &\scalebox{0.78}{0.270}&{\scalebox{0.78}{0.176}} & {\scalebox{0.78}{0.262}}& {\scalebox{0.78}{0.180}} & {\scalebox{0.78}{0.264}} & \scalebox{0.78}{0.182} & \scalebox{0.78}{0.265} & {{\scalebox{0.78}{0.175}}} & \secondres{{\scalebox{0.78}{0.259}}} & \scalebox{0.78}{0.287} & \scalebox{0.78}{0.366} & \scalebox{0.78}{0.207} & \scalebox{0.78}{0.305} &{\scalebox{0.78}{0.187}} &\scalebox{0.78}{0.267} &\scalebox{0.78}{0.193} &\scalebox{0.78}{0.292} & \scalebox{0.78}{0.286} & \scalebox{0.78}{0.377} &\scalebox{0.78}{0.203} &\scalebox{0.78}{0.287} &{\scalebox{0.78}{0.192}} &\scalebox{0.78}{0.274} &\scalebox{0.78}{0.255} &\scalebox{0.78}{0.339} \\ 
    & \scalebox{0.78}{192} & \boldres{\scalebox{0.78}{0.238}} & \secondres{\scalebox{0.78}{0.301}}
    &\secondres{\scalebox{0.78}{0.242}} &\boldres{\scalebox{0.78}{0.298}}
    &\scalebox{0.78}{0.247} &\scalebox{0.78}{0.310}
    &\scalebox{0.78}{0.248} &\scalebox{0.78}{0.308}& {\scalebox{0.78}{0.244}} &{\scalebox{0.78}{0.306}} & {\scalebox{0.78}{0.250}} &{\scalebox{0.78}{0.309}} & {\scalebox{0.78}{0.246}} & {\scalebox{0.78}{0.304}} & {\scalebox{0.78}{0.241}} & {\scalebox{0.78}{0.302}} & \scalebox{0.78}{0.414} & \scalebox{0.78}{0.492} & \scalebox{0.78}{0.290} & \scalebox{0.78}{0.364} &{\scalebox{0.78}{0.249}} &{\scalebox{0.78}{0.309}} &\scalebox{0.78}{0.284} &\scalebox{0.78}{0.362} & \scalebox{0.78}{0.399} & \scalebox{0.78}{0.445} &\scalebox{0.78}{0.269} &\scalebox{0.78}{0.328} &\scalebox{0.78}{0.280} &\scalebox{0.78}{0.339} &\scalebox{0.78}{0.281} &\scalebox{0.78}{0.340} \\ 
    
    & \scalebox{0.78}{336} & \secondres{\scalebox{0.78}{0.300}} & \secondres{\scalebox{0.78}{0.340}}&
    \boldres{\scalebox{0.78}{0.297}} &\boldres{\scalebox{0.78}{0.332}}&
    {\scalebox{0.78}{0.307}} &{\scalebox{0.78}{0.349}}&
    \scalebox{0.78}{0.309} &\scalebox{0.78}{0.348}& {\scalebox{0.78}{0.309}} & {\scalebox{0.78}{0.348}}& {\scalebox{0.78}{0.311}} & {\scalebox{0.78}{0.348}} & {\scalebox{0.78}{0.307}} & {\scalebox{0.78}{0.342}} & {\scalebox{0.78}{0.305}} & {\scalebox{0.78}{0.343}}  & \scalebox{0.78}{0.597} & \scalebox{0.78}{0.542}  & \scalebox{0.78}{0.377} & \scalebox{0.78}{0.422} &{\scalebox{0.78}{0.321}} &{\scalebox{0.78}{0.351}} &\scalebox{0.78}{0.369} &\scalebox{0.78}{0.427} & \scalebox{0.78}{0.637} & \scalebox{0.78}{0.591} &\scalebox{0.78}{0.325} &\scalebox{0.78}{0.366} &\scalebox{0.78}{0.334} &\scalebox{0.78}{0.361} &\scalebox{0.78}{0.339} &\scalebox{0.78}{0.372} \\ 
    
    & \scalebox{0.78}{720} & \secondres{\scalebox{0.78}{0.394}} & \boldres{\scalebox{0.78}{0.394}}
    &\boldres{\scalebox{0.78}{0.389}} &\scalebox{0.78}{0.404}
    &{\scalebox{0.78}{0.400}} &\scalebox{0.78}{0.408}
    &\scalebox{0.78}{0.411} &\scalebox{0.78}{0.402}& \scalebox{0.78}{0.412} & \scalebox{0.78}{0.410}& \scalebox{0.78}{0.412} & \scalebox{0.78}{0.407} & {\scalebox{0.78}{0.407}} & {{\scalebox{0.78}{0.398}}} & {{\scalebox{0.78}{0.402}}} & \secondres{\scalebox{0.78}{0.400}} & \scalebox{0.78}{1.730} & \scalebox{0.78}{1.042} & \scalebox{0.78}{0.558} & \scalebox{0.78}{0.524} &{\scalebox{0.78}{0.408}} &{\scalebox{0.78}{0.403}} &\scalebox{0.78}{0.554} &\scalebox{0.78}{0.522} & \scalebox{0.78}{0.960} & \scalebox{0.78}{0.735} &\scalebox{0.78}{0.421} &\scalebox{0.78}{0.415} &\scalebox{0.78}{0.417} &\scalebox{0.78}{0.413} &\scalebox{0.78}{0.433} &\scalebox{0.78}{0.432} \\ 
    \cmidrule(lr){2-34}
    & \scalebox{0.78}{Avg} & \secondres{\scalebox{0.78}{0.276}} & \boldres{\scalebox{0.78}{0.322}} 
    &\boldres{\scalebox{0.78}{0.275}} & \boldres{\scalebox{0.78}{0.322}}
    &\scalebox{0.78}{0.283} & \scalebox{0.78}{0.335}
    &\scalebox{0.78}{0.288} & \scalebox{0.78}{0.332} &\scalebox{0.78}{0.285} & \scalebox{0.78}{0.331}& {\scalebox{0.78}{0.288}} & {\scalebox{0.78}{0.332}} & {\scalebox{0.78}{0.286}} & {\scalebox{0.78}{0.327}} & {\scalebox{0.78}{0.281}} & \secondres{\scalebox{0.78}{0.326}} & \scalebox{0.78}{0.757} & \scalebox{0.78}{0.610} & \scalebox{0.78}{0.358} & \scalebox{0.78}{0.404} &{\scalebox{0.78}{0.291}} &{\scalebox{0.78}{0.333}} &\scalebox{0.78}{0.350} &\scalebox{0.78}{0.401} & \scalebox{0.78}{0.571} & \scalebox{0.78}{0.537} &\scalebox{0.78}{0.305} &\scalebox{0.78}{0.349} &\scalebox{0.78}{0.306} &\scalebox{0.78}{0.347} &\scalebox{0.78}{0.327} &\scalebox{0.78}{0.371} \\ 
    \midrule
    
    \multirow{4}{*}{\rotatebox{90}{{\scalebox{0.95}{ETTh1}}}}
    &  \scalebox{0.78}{96} & \secondres{{\scalebox{0.78}{0.376}}} & {{\scalebox{0.78}{0.400}}}
    &\secondres{\scalebox{0.78}{0.376}} &\scalebox{0.78}{0.404}
    &\boldres{\scalebox{0.78}{0.373}} &\secondres{\scalebox{0.78}{0.396}}
    &\scalebox{0.78}{0.444} &\scalebox{0.78}{0.439}& {\scalebox{0.78}{0.380}} & {\scalebox{0.78}{0.396}}& {\scalebox{0.78}{0.386}} & {\scalebox{0.78}{0.405}} & \scalebox{0.78}{0.386} & \boldres{{\scalebox{0.78}{0.395}}} & \scalebox{0.78}{0.414} & \scalebox{0.78}{0.419} & \scalebox{0.78}{0.423} & \scalebox{0.78}{0.448} & \scalebox{0.78}{0.479}& \scalebox{0.78}{0.464}  &{{\scalebox{0.78}{0.384}}} &{\scalebox{0.78}{0.402}} & \scalebox{0.78}{0.386} &{{\scalebox{0.78}{0.400}}} & \scalebox{0.78}{0.654} & \scalebox{0.78}{0.599} &\secondres{{\scalebox{0.78}{0.376}}} &\scalebox{0.78}{0.419} &\scalebox{0.78}{0.513} &\scalebox{0.78}{0.491} &\scalebox{0.78}{0.449} &\scalebox{0.78}{0.459}  \\ 
    & \scalebox{0.78}{192} 
    & {\scalebox{0.78}{0.432}} & \secondres{\scalebox{0.78}{0.429}}
    &\scalebox{0.78}{0.424} &\scalebox{0.78}{0.433}
    &\secondres{\scalebox{0.78}{0.422}} &\boldres{\scalebox{0.78}{0.424}}
    &\scalebox{0.78}{0.501} &\scalebox{0.78}{0.471}& {\scalebox{0.78}{0.432}} & \scalebox{0.78}{0.427}& \scalebox{0.78}{0.441} & \scalebox{0.78}{0.436} & {\scalebox{0.78}{0.437}} & \boldres{\scalebox{0.78}{0.424}} & \scalebox{0.78}{0.460} & \scalebox{0.78}{0.445} & \scalebox{0.78}{0.471} & \scalebox{0.78}{0.474}  & \scalebox{0.78}{0.525} & \scalebox{0.78}{0.492} &{\scalebox{0.78}{0.436}} &\secondres{\scalebox{0.78}{0.429}}  &{\scalebox{0.78}{0.437}} &{\scalebox{0.78}{0.432}} & \scalebox{0.78}{0.719} & \scalebox{0.78}{0.631} &\boldres{\scalebox{0.78}{0.420}} &\scalebox{0.78}{0.448} &\scalebox{0.78}{0.534} &\scalebox{0.78}{0.504} &\scalebox{0.78}{0.500} &\scalebox{0.78}{0.482} \\ 
    & \scalebox{0.78}{336} 
    & {\scalebox{0.78}{0.477}} & \boldres{\scalebox{0.78}{0.437}}
     &\boldres{\scalebox{0.78}{0.455}} &\secondres{\scalebox{0.78}{0.441}}
      &\boldres{\scalebox{0.78}{0.455}} &\scalebox{0.78}{0.442}
    &\scalebox{0.78}{0.529} &\scalebox{0.78}{0.485}& {\scalebox{0.78}{0.480}} & \boldres{\scalebox{0.78}{0.437}}& {\scalebox{0.78}{0.487}} & {\scalebox{0.78}{0.458}} & {\scalebox{0.78}{0.479}} & {\scalebox{0.78}{0.446}} & \scalebox{0.78}{0.501} & \scalebox{0.78}{0.466} & \scalebox{0.78}{0.570} & \scalebox{0.78}{0.546} & \scalebox{0.78}{0.565} & \scalebox{0.78}{0.515} &\scalebox{0.78}{0.491} &\scalebox{0.78}{0.469} &{\scalebox{0.78}{0.481}} & {\scalebox{0.78}{0.459}} & \scalebox{0.78}{0.778} & \scalebox{0.78}{0.659} &\secondres{\scalebox{0.78}{0.459}} &{\scalebox{0.78}{0.465}} &\scalebox{0.78}{0.588} &\scalebox{0.78}{0.535} &\scalebox{0.78}{0.521} &\scalebox{0.78}{0.496} \\ 
    & \scalebox{0.78}{720} 
    & {\scalebox{0.78}{0.488}} & {\scalebox{0.78}{0.471}}
     &\secondres{\scalebox{0.78}{0.458}} &\secondres{\scalebox{0.78}{0.463}}
      &\boldres{\scalebox{0.78}{0.455}} &\boldres{\scalebox{0.78}{0.461}}
    &\scalebox{0.78}{0.524} &\scalebox{0.78}{0.501} & {\scalebox{0.78}{0.504}} & {\scalebox{0.78}{0.485}} & {\scalebox{0.78}{0.503}} & {\scalebox{0.78}{0.491}} & {\scalebox{0.78}{0.481}} & 
    {\scalebox{0.78}{0.470}} & {\scalebox{0.78}{0.500}} & {\scalebox{0.78}{0.488}} & \scalebox{0.78}{0.653} & \scalebox{0.78}{0.621} & \scalebox{0.78}{0.594} & \scalebox{0.78}{0.558} &\scalebox{0.78}{0.521} &{\scalebox{0.78}{0.500}} &\scalebox{0.78}{0.519} &\scalebox{0.78}{0.516} & \scalebox{0.78}{0.836} & \scalebox{0.78}{0.699} &{\scalebox{0.78}{0.506}} &{\scalebox{0.78}{0.507}} &\scalebox{0.78}{0.643} &\scalebox{0.78}{0.616} &{\scalebox{0.78}{0.514}} &\scalebox{0.78}{0.512}  \\ 
    \cmidrule(lr){2-34}
    & \scalebox{0.78}{Avg} 
    &{\scalebox{0.78}{0.443}} & \secondres{\scalebox{0.78}{0.432}}
    &\secondres{\scalebox{0.78}{0.428}} & \scalebox{0.78}{0.435}
    &\boldres{\scalebox{0.78}{0.426}} & \boldres{\scalebox{0.78}{0.431}}
    &\scalebox{0.78}{0.500} & \scalebox{0.78}{0.474}&
    \scalebox{0.78}{0.450} & \scalebox{0.78}{0.439}& {\scalebox{0.78}{0.454}} & {\scalebox{0.78}{0.447}} & {\scalebox{0.78}{0.446}} & {\scalebox{0.78}{0.434}} & \scalebox{0.78}{0.469} & \scalebox{0.78}{0.454} & \scalebox{0.78}{0.529} & \scalebox{0.78}{0.522} & \scalebox{0.78}{0.541} & \scalebox{0.78}{0.507} &\scalebox{0.78}{0.458} &{\scalebox{0.78}{0.450}} &{\scalebox{0.78}{0.456}} &{\scalebox{0.78}{0.452}} & \scalebox{0.78}{0.747} & \scalebox{0.78}{0.647} &{\scalebox{0.78}{0.440}} &\scalebox{0.78}{0.460} &\scalebox{0.78}{0.570} &\scalebox{0.78}{0.537} &\scalebox{0.78}{0.496} &\scalebox{0.78}{0.487}  \\ 
    \midrule

    \multirow{4}{*}{\rotatebox{90}{\scalebox{0.95}{ETTh2}}}
    &  \scalebox{0.78}{96} 
    & \secondres{\scalebox{0.78}{0.288}} & {\scalebox{0.78}{0.340}}
    &\boldres{\scalebox{0.78}{0.280}} &\boldres{\scalebox{0.78}{0.337}}
    &\scalebox{0.78}{0.297} &\scalebox{0.78}{0.357}
    
    &\scalebox{0.78}{0.318} &\scalebox{0.78}{0.361}& {\scalebox{0.78}{0.305}} & {\scalebox{0.78}{0.354}}& {\scalebox{0.78}{0.297}} & {\scalebox{0.78}{0.349}} & \secondres{\scalebox{0.78}{0.288}} & \secondres{\scalebox{0.78}{0.338}} & {\scalebox{0.78}{0.302}} & {\scalebox{0.78}{0.348}} & \scalebox{0.78}{0.745} & \scalebox{0.78}{0.584} &\scalebox{0.78}{0.400} & \scalebox{0.78}{0.440}  & {\scalebox{0.78}{0.340}} & {\scalebox{0.78}{0.374}} &{\scalebox{0.78}{0.333}} &{\scalebox{0.78}{0.387}} & \scalebox{0.78}{0.707} & \scalebox{0.78}{0.621}  &\scalebox{0.78}{0.358} &\scalebox{0.78}{0.397} &\scalebox{0.78}{0.476} &\scalebox{0.78}{0.458} &\scalebox{0.78}{0.346} &\scalebox{0.78}{0.388} \\ 
    
    & \scalebox{0.78}{192} 
    & \secondres{\scalebox{0.78}{0.373}} & \boldres{\scalebox{0.78}{0.390}}
     &\boldres{\scalebox{0.78}{0.366}} &\secondres{\scalebox{0.78}{0.391}} 
      &\secondres{\scalebox{0.78}{0.373}} &\scalebox{0.78}{0.400} 
    &\scalebox{0.78}{0.399} &\scalebox{0.78}{0.409} & {\scalebox{0.78}{0.389}} & {\scalebox{0.78}{0.408}} & {\scalebox{0.78}{0.380}} & {\scalebox{0.78}{0.400}} &{\scalebox{0.78}{0.374}} & \boldres{\scalebox{0.78}{0.390}} &{\scalebox{0.78}{0.388}} & {\scalebox{0.78}{0.400}} & \scalebox{0.78}{0.877} & \scalebox{0.78}{0.656} & \scalebox{0.78}{0.528} & \scalebox{0.78}{0.509} & {\scalebox{0.78}{0.402}} & {\scalebox{0.78}{0.414}} &\scalebox{0.78}{0.477} &\scalebox{0.78}{0.476} & \scalebox{0.78}{0.860} & \scalebox{0.78}{0.689} &{\scalebox{0.78}{0.429}} &{\scalebox{0.78}{0.439}} &\scalebox{0.78}{0.512} &\scalebox{0.78}{0.493} &\scalebox{0.78}{0.456} &\scalebox{0.78}{0.452} \\ 
    & \scalebox{0.78}{336} 
    & \boldres{\scalebox{0.78}{0.380}} & \boldres{\scalebox{0.78}{0.406}}
    &\secondres{\scalebox{0.78}{0.415}} &\scalebox{0.78}{0.429}
    &\scalebox{0.78}{0.421} &\scalebox{0.78}{0.440}
    &\scalebox{0.78}{0.435} &\scalebox{0.78}{0.441}& {\scalebox{0.78}{0.415}} & {\scalebox{0.78}{0.432}}& {\scalebox{0.78}{0.428}} & {\scalebox{0.78}{0.432}} & \secondres{\scalebox{0.78}{0.415}} & \secondres{\scalebox{0.78}{0.426}} & {\scalebox{0.78}{0.426}} & {\scalebox{0.78}{0.433}}& \scalebox{0.78}{1.043} & \scalebox{0.78}{0.731} & \scalebox{0.78}{0.643} & \scalebox{0.78}{0.571}  & {\scalebox{0.78}{0.452}} & {\scalebox{0.78}{0.452}} &\scalebox{0.78}{0.594} &\scalebox{0.78}{0.541} & \scalebox{0.78}{1.000} &\scalebox{0.78}{0.744} &\scalebox{0.78}{0.496} &\scalebox{0.78}{0.487} &\scalebox{0.78}{0.552} &\scalebox{0.78}{0.551} &{\scalebox{0.78}{0.482}} &\scalebox{0.78}{0.486}\\ 
    & \scalebox{0.78}{720} 
    & \boldres{\scalebox{0.78}{0.412}} & \boldres{\scalebox{0.78}{0.432}}
    &\scalebox{0.78}{0.425} &\scalebox{0.78}{0.446}
    &\scalebox{0.78}{0.441} &\scalebox{0.78}{0.467}
    &\scalebox{0.78}{0.438} &\scalebox{0.78}{0.451}& {\scalebox{0.78}{0.432}} & {\scalebox{0.78}{0.451}}& {\scalebox{0.78}{0.427}} & {\scalebox{0.78}{0.445}} & \secondres{\scalebox{0.78}{0.420}} & \secondres{\scalebox{0.78}{0.440}} & {\scalebox{0.78}{0.431}} & {\scalebox{0.78}{0.446}} & \scalebox{0.78}{1.104} & \scalebox{0.78}{0.763} & \scalebox{0.78}{0.874} & \scalebox{0.78}{0.679} & {\scalebox{0.78}{0.462}} & {\scalebox{0.78}{0.468}} &\scalebox{0.78}{0.831} &\scalebox{0.78}{0.657} & \scalebox{0.78}{1.249} & \scalebox{0.78}{0.838} &{\scalebox{0.78}{0.463}} &{\scalebox{0.78}{0.474}} &\scalebox{0.78}{0.562} &\scalebox{0.78}{0.560} &\scalebox{0.78}{0.515} &\scalebox{0.78}{0.511} \\ 

    \cmidrule(lr){2-34}
    & \scalebox{0.78}{Avg} 
    & \boldres{\scalebox{0.78}{0.363}} & \boldres{\scalebox{0.78}{0.392}}
    &\secondres{\scalebox{0.78}{0.372}} & \scalebox{0.78}{0.401}
    &\scalebox{0.78}{0.383} & \scalebox{0.78}{0.416}
    &\scalebox{0.78}{0.398} & \scalebox{0.78}{0.416}
    &\scalebox{0.78}{0.385} & \scalebox{0.78}{0.411}
    & {\scalebox{0.78}{0.383}} & {\scalebox{0.78}{0.407}} & {\scalebox{0.78}{0.374}} & \secondres{\scalebox{0.78}{0.398}} & {\scalebox{0.78}{0.387}} & {\scalebox{0.78}{0.407}} & \scalebox{0.78}{0.942} & \scalebox{0.78}{0.684} & \scalebox{0.78}{0.611} & \scalebox{0.78}{0.550}  &{\scalebox{0.78}{0.414}} &{\scalebox{0.78}{0.427}} &\scalebox{0.78}{0.559} &\scalebox{0.78}{0.515} & \scalebox{0.78}{0.954} & \scalebox{0.78}{0.723} &\scalebox{0.78}{{0.437}} &\scalebox{0.78}{{0.449}} &\scalebox{0.78}{0.526} &\scalebox{0.78}{0.516} &\scalebox{0.78}{0.450} &\scalebox{0.78}{0.459} \\ 
    \midrule
    
    \multirow{4}{*}{\rotatebox{90}{\scalebox{0.95}{ECL}}} 
    &  \scalebox{0.78}{96} 
    & \secondres{\scalebox{0.78}{0.146}} & \secondres{\scalebox{0.78}{0.244}}
    &\boldres{\scalebox{0.78}{0.145}} &\scalebox{0.78}{0.245}
    &\scalebox{0.78}{0.162} &\scalebox{0.78}{0.255}
    &\scalebox{0.78}{0.181} &\scalebox{0.78}{0.271}& {\scalebox{0.78}{0.190}} & {\scalebox{0.78}{0.273}}& {\scalebox{0.78}{0.148}} & \boldres{\scalebox{0.78}{0.240}} & \scalebox{0.78}{0.201} & \scalebox{0.78}{0.281} & \scalebox{0.78}{0.181} & {\scalebox{0.78}{0.270}} & \scalebox{0.78}{0.219} & \scalebox{0.78}{0.314} & \scalebox{0.78}{0.237} & \scalebox{0.78}{0.329} &{\scalebox{0.78}{0.168}} &\scalebox{0.78}{0.272} &\scalebox{0.78}{0.197} &\scalebox{0.78}{0.282} & \scalebox{0.78}{0.247} & \scalebox{0.78}{0.345} &\scalebox{0.78}{0.193} &\scalebox{0.78}{0.308} &{\scalebox{0.78}{0.169}} &{\scalebox{0.78}{0.273}} &\scalebox{0.78}{0.201} &\scalebox{0.78}{0.317}  \\ 
    & \scalebox{0.78}{192} 
    & \boldres{{\scalebox{0.78}{0.162}}} & \secondres{{\scalebox{0.78}{0.258}}}
    &\secondres{\scalebox{0.78}{0.166}} &\scalebox{0.78}{0.263}
    &{\scalebox{0.78}{0.174}} &\scalebox{0.78}{0.267}
    &{\scalebox{0.78}{0.188}} &\scalebox{0.78}{0.277}& {\scalebox{0.78}{0.191}} & {\scalebox{0.78}{0.274}} & \boldres{\scalebox{0.78}{0.162}} & \boldres{\scalebox{0.78}{0.253}} & \scalebox{0.78}{0.201} & \scalebox{0.78}{0.283} & \scalebox{0.78}{0.188} & {\scalebox{0.78}{0.274}} & \scalebox{0.78}{0.231} & \scalebox{0.78}{0.322} & \scalebox{0.78}{0.236} & \scalebox{0.78}{0.330} &{\scalebox{0.78}{0.184}} &\scalebox{0.78}{0.289} &\scalebox{0.78}{0.196} &{\scalebox{0.78}{0.285}} & \scalebox{0.78}{0.257} & \scalebox{0.78}{0.355} &\scalebox{0.78}{0.201} &\scalebox{0.78}{0.315} &{\scalebox{0.78}{0.182}} &\scalebox{0.78}{0.286} &\scalebox{0.78}{0.222} &\scalebox{0.78}{0.334} \\ 
    & \scalebox{0.78}{336} 
    & \secondres{\scalebox{0.78}{0.177}} & {\scalebox{0.78}{0.274}}
    &\boldres{\scalebox{0.78}{0.172}} &\secondres{\scalebox{0.78}{0.270}}
    &\scalebox{0.78}{0.191} &\scalebox{0.78}{0.285}
    
    &\scalebox{0.78}{0.204} &\scalebox{0.78}{0.292}& {\scalebox{0.78}{0.208}} & {\scalebox{0.78}{0.294}}& {\scalebox{0.78}{0.178}} & \boldres{\scalebox{0.78}{0.269}} & \scalebox{0.78}{0.215} & \scalebox{0.78}{0.298} & \scalebox{0.78}{0.204} & {\scalebox{0.78}{0.293}} & \scalebox{0.78}{0.246} & \scalebox{0.78}{0.337} & \scalebox{0.78}{0.249} & \scalebox{0.78}{0.344} &{\scalebox{0.78}{0.198}} &{\scalebox{0.78}{0.300}} &\scalebox{0.78}{0.209} &{\scalebox{0.78}{0.301}} & \scalebox{0.78}{0.269} & \scalebox{0.78}{0.369} &\scalebox{0.78}{0.214} &\scalebox{0.78}{0.329} &{\scalebox{0.78}{0.200}} &\scalebox{0.78}{0.304} &\scalebox{0.78}{0.231} &\scalebox{0.78}{0.338}  \\ 
    & \scalebox{0.78}{720} 
    & \boldres{\scalebox{0.78}{0.202}} & \boldres{\scalebox{0.78}{0.297}}
    &\secondres{\scalebox{0.78}{0.209}} &\secondres{\scalebox{0.78}{0.307}}
    &\scalebox{0.78}{0.230} &\scalebox{0.78}{0.321}
    &\scalebox{0.78}{0.244} &\scalebox{0.78}{0.322}& {\scalebox{0.78}{0.251}} & {\scalebox{0.78}{0.326}}& {\scalebox{0.78}{0.225}} & {\scalebox{0.78}{0.317}} & \scalebox{0.78}{0.257} & \scalebox{0.78}{0.331} & \scalebox{0.78}{0.246} & \scalebox{0.78}{0.324} & \scalebox{0.78}{0.280} & \scalebox{0.78}{0.363} & \scalebox{0.78}{0.284} & \scalebox{0.78}{0.373} &{\scalebox{0.78}{0.220}} &{\scalebox{0.78}{0.320}} &\scalebox{0.78}{0.245} &\scalebox{0.78}{0.333} & \scalebox{0.78}{0.299} & \scalebox{0.78}{0.390} &\scalebox{0.78}{0.246} &\scalebox{0.78}{0.355} &{\scalebox{0.78}{0.222}} &{\scalebox{0.78}{0.321}} &\scalebox{0.78}{0.254} &\scalebox{0.78}{0.361} \\ 
    \cmidrule(lr){2-34}
    & \scalebox{0.78}{Avg} & \boldres{\scalebox{0.78}{0.172}} & \boldres{\scalebox{0.78}{0.268}}
    &\secondres{\scalebox{0.78}{0.173}} & \scalebox{0.78}{0.271}
    &\scalebox{0.78}{0.189} & \scalebox{0.78}{0.282}
    &\scalebox{0.78}{0.204} & \scalebox{0.78}{0.291}&\scalebox{0.78}{0.210} & \scalebox{0.78}{0.291}& {\scalebox{0.78}{0.178}} & \secondres{\scalebox{0.78}{0.270}} & \scalebox{0.78}{0.219} & \scalebox{0.78}{0.298} & \scalebox{0.78}{0.205} & {\scalebox{0.78}{0.290}} & \scalebox{0.78}{0.244} & \scalebox{0.78}{0.334} & \scalebox{0.78}{0.251} & \scalebox{0.78}{0.344} &{\scalebox{0.78}{0.192}} &\scalebox{0.78}{0.295} &\scalebox{0.78}{0.212} &\scalebox{0.78}{0.300} & \scalebox{0.78}{0.268} & \scalebox{0.78}{0.365} &\scalebox{0.78}{0.214} &\scalebox{0.78}{0.327} &{\scalebox{0.78}{0.193}} &{\scalebox{0.78}{0.296}} &\scalebox{0.78}{0.227} &\scalebox{0.78}{0.338} \\ 

        \midrule
    
    \multirow{5}{*}{\rotatebox{90}{{\scalebox{0.95}{Exchange}}}}
    &  \scalebox{0.78}{96} 
    &\boldres{\scalebox{0.78}{0.083}} & {\scalebox{0.78}{0.202}}
&\scalebox{0.78}{0.086} & \scalebox{0.78}{0.205}
    &\secondres{\scalebox{0.78}{0.084}} & {\scalebox{0.78}{0.201}}
    & \boldres{\scalebox{0.78}{0.083}} &\boldres{{\scalebox{0.78}{0.198}}}
    & \boldres{\scalebox{0.78}{0.083}} & \secondres{{\scalebox{0.78}{0.200}}}
    & {\scalebox{0.78}{0.086}} & {\scalebox{0.78}{0.206}} & \scalebox{0.78}{0.093} & \scalebox{0.78}{0.217} & {\scalebox{0.78}{0.088}} & {\scalebox{0.78}{0.205}} & \scalebox{0.78}{0.256} & \scalebox{0.78}{0.367} & \scalebox{0.78}{0.094} & \scalebox{0.78}{0.218} & \scalebox{0.78}{0.107} & \scalebox{0.78}{0.234} & \scalebox{0.78}{0.088} & \scalebox{0.78}{0.218} & \scalebox{0.78}{0.267} & \scalebox{0.78}{0.396} & \scalebox{0.78}{0.148} & \scalebox{0.78}{0.278} & \scalebox{0.78}{0.111} & \scalebox{0.78}{0.237} & \scalebox{0.78}{0.197} & \scalebox{0.78}{0.323} \\ 

    &  \scalebox{0.78}{192}
    &{\scalebox{0.78}{0.176}} & {\scalebox{0.78}{0.298}}
    &\secondres{\scalebox{0.78}{0.175}} & \boldres{\scalebox{0.78}{0295}}
    &\scalebox{0.78}{0.176} & \secondres{\scalebox{0.78}{0.296}}
    &\scalebox{0.78}{0.182} & \scalebox{0.78}{0.303}
   &\boldres{\scalebox{0.78}{0.174}} & \boldres{\scalebox{0.78}{0.295}}
    & \scalebox{0.78}{0.177} & {\scalebox{0.78}{0.299}} & \scalebox{0.78}{0.184} & \scalebox{0.78}{0.307} & {\scalebox{0.78}{0.176}} & {\scalebox{0.78}{0.299}} & \scalebox{0.78}{0.470} & \scalebox{0.78}{0.509} & \scalebox{0.78}{0.184} & \scalebox{0.78}{0.307} & \scalebox{0.78}{0.226} & \scalebox{0.78}{0.344} & {\scalebox{0.78}{0.176}} & \scalebox{0.78}{0.315} & \scalebox{0.78}{0.351} & \scalebox{0.78}{0.459} & \scalebox{0.78}{0.271} & \scalebox{0.78}{0.315} & \scalebox{0.78}{0.219} & \scalebox{0.78}{0.335} & \scalebox{0.78}{0.300} & \scalebox{0.78}{0.369} \\ 

    &  \scalebox{0.78}{336} 
    &\scalebox{0.78}{0.327} & \secondres{\scalebox{0.78}{0.413}}
     &\scalebox{0.78}{0.330} & \scalebox{0.78}{0.414}
      &\scalebox{0.78}{0.332} & \scalebox{0.78}{0.416}
     &\scalebox{0.78}{0.342} & \scalebox{0.78}{0.432}
    &\scalebox{0.78}{0.349} & \scalebox{0.78}{0.427}
    & \scalebox{0.78}{0.331} & {\scalebox{0.78}{0.417}} & \scalebox{0.78}{0.351} & \scalebox{0.78}{0.432}& \boldres{\scalebox{0.78}{0.301}} & \boldres{\scalebox{0.78}{0.397}} & \scalebox{0.78}{1.268} & \scalebox{0.78}{0.883} & \scalebox{0.78}{0.349} & \scalebox{0.78}{0.431} & \scalebox{0.78}{0.367} & \scalebox{0.78}{0.448} & \secondres{\scalebox{0.78}{0.313}} & \scalebox{0.78}{0.427} & \scalebox{0.78}{1.324} & \scalebox{0.78}{0.853} & \scalebox{0.78}{0.460} & \scalebox{0.78}{0.427} & \scalebox{0.78}{0.421} & \scalebox{0.78}{0.476} & \scalebox{0.78}{0.509} & \scalebox{0.78}{0.524} \\ 
    &  \scalebox{0.78}{720} 
    &\boldres{\scalebox{0.78}{0.839}} & \boldres{\scalebox{0.78}{0.689}}
    &\scalebox{0.78}{0.927} & \scalebox{0.78}{0.736}
    &\scalebox{0.78}{0.935} & \scalebox{0.78}{0.736}
    &\scalebox{0.78}{0.970} & \scalebox{0.78}{0.745}
    &\scalebox{0.78}{0.864} & \scalebox{0.78}{0.700}
    & \secondres{\scalebox{0.78}{0.847}} & \secondres{\scalebox{0.78}{0.691}} & \scalebox{0.78}{0.886} & \scalebox{0.78}{0.714} & \scalebox{0.78}{0.901} & \scalebox{0.78}{0.714} & \scalebox{0.78}{1.767} & \scalebox{0.78}{1.068} & \scalebox{0.78}{0.852} & \scalebox{0.78}{0.698} & \scalebox{0.78}{0.964} & \scalebox{0.78}{0.746} & \boldres{\scalebox{0.78}{0.839}} & \scalebox{0.78}{0.695} & \scalebox{0.78}{1.058} & \scalebox{0.78}{0.797} & \scalebox{0.78}{1.195} & {\scalebox{0.78}{0.695}} & \scalebox{0.78}{1.092} & \scalebox{0.78}{0.769} & \scalebox{0.78}{1.447} & \scalebox{0.78}{0.941} \\ 
    \cmidrule(lr){2-34}
    &  \scalebox{0.78}{Avg} 
    &\secondres{\scalebox{0.78}{0.356}} & \boldres{\scalebox{0.78}{0.401}}
     &\scalebox{0.78}{0.378} & \scalebox{0.78}{0.411}
      &\scalebox{0.78}{0.382} & \scalebox{0.78}{0.412}
    &\scalebox{0.78}{0.395} & \scalebox{0.78}{0.421}
    &\scalebox{0.78}{0.368} & \scalebox{0.78}{0.406}
    & {\scalebox{0.78}{0.360}} & \secondres{\scalebox{0.78}{0.403}} & \scalebox{0.78}{0.378} & \scalebox{0.78}{0.417} & \scalebox{0.78}{0.367} & {\scalebox{0.78}{0.404}} & \scalebox{0.78}{0.940} & \scalebox{0.78}{0.707} & \scalebox{0.78}{0.370} & \scalebox{0.78}{0.413} & \scalebox{0.78}{0.416} & \scalebox{0.78}{0.443} & \boldres{\scalebox{0.78}{0.354}} & \scalebox{0.78}{0.414} & \scalebox{0.78}{0.750} & \scalebox{0.78}{0.626} & \scalebox{0.78}{0.519} & \scalebox{0.78}{0.429} & \scalebox{0.78}{0.461} & \scalebox{0.78}{0.454} & \scalebox{0.78}{0.613} & \scalebox{0.78}{0.539} \\ 

    \midrule
    
    \multirow{4}{*}{\rotatebox{90}{\scalebox{0.95}{Traffic}}} 
    & \scalebox{0.78}{96} 
    & \boldres{\scalebox{0.78}{0.388}} & \boldres{\scalebox{0.78}{0.261}} 
    &\scalebox{0.78}{0.600} &\scalebox{0.78}{0.310}
    &\scalebox{0.78}{0.651} &\scalebox{0.78}{0.327}
    &\scalebox{0.78}{0.455} &\scalebox{0.78}{0.313}& {\scalebox{0.78}{0.498}} & {\scalebox{0.78}{0.330}} & \secondres{\scalebox{0.78}{0.395}} & \secondres{\scalebox{0.78}{0.268}} & \scalebox{0.78}{0.649} & \scalebox{0.78}{0.389} & {\scalebox{0.78}{0.462}} & \scalebox{0.78}{0.295} & \scalebox{0.78}{0.522} & {\scalebox{0.78}{0.290}} & \scalebox{0.78}{0.805} & \scalebox{0.78}{0.493} &{\scalebox{0.78}{0.593}} &{\scalebox{0.78}{0.321}} &\scalebox{0.78}{0.650} &\scalebox{0.78}{0.396} & \scalebox{0.78}{0.788} & \scalebox{0.78}{0.499} &{\scalebox{0.78}{0.587}} &\scalebox{0.78}{0.366} &\scalebox{0.78}{0.612} &{\scalebox{0.78}{0.338}} &\scalebox{0.78}{0.613} &\scalebox{0.78}{0.388} \\ 
    & \scalebox{0.78}{192} 
    & \boldres{\scalebox{0.78}{0.411}} & \boldres{\scalebox{0.78}{0.271}}
     &\scalebox{0.78}{0.609} &\scalebox{0.78}{0.309}
      &\scalebox{0.78}{0.675} &\scalebox{0.78}{0.332}
    &\scalebox{0.78}{0.469} &\scalebox{0.78}{0.315}& {\scalebox{0.78}{0.500}} & {\scalebox{0.78}{0.332}}& \secondres{{\scalebox{0.78}{0.417}}} & \secondres{{\scalebox{0.78}{0.276}}} & \scalebox{0.78}{0.601} & \scalebox{0.78}{0.366} & {\scalebox{0.78}{0.466}} & \scalebox{0.78}{0.296} & \scalebox{0.78}{0.530} & {\scalebox{0.78}{0.293}} & \scalebox{0.78}{0.756} & \scalebox{0.78}{0.474} &\scalebox{0.78}{0.617} &{\scalebox{0.78}{0.336}} &{\scalebox{0.78}{0.598}} &\scalebox{0.78}{0.370} & \scalebox{0.78}{0.789} & \scalebox{0.78}{0.505} &\scalebox{0.78}{0.604} &\scalebox{0.78}{0.373} &\scalebox{0.78}{0.613} &{\scalebox{0.78}{0.340}} &\scalebox{0.78}{0.616} &\scalebox{0.78}{0.382}  \\ 
    & \scalebox{0.78}{336} 
    & \boldres{\scalebox{0.78}{0.428}} & \boldres{\scalebox{0.78}{0.278}}
    &\scalebox{0.78}{0.622} &\scalebox{0.78}{0.300}
    &\scalebox{0.78}{0.678} &\scalebox{0.78}{0.341}
    &\scalebox{0.78}{0.482} &\scalebox{0.78}{0.319}& {\scalebox{0.78}{0.512}} & {\scalebox{0.78}{0.332}}& \secondres{\scalebox{0.78}{0.433}} & \secondres{\scalebox{0.78}{0.283}} & \scalebox{0.78}{0.609} & \scalebox{0.78}{0.369} & {\scalebox{0.78}{0.482}} & {\scalebox{0.78}{0.304}} & \scalebox{0.78}{0.558} & \scalebox{0.78}{0.305}  & \scalebox{0.78}{0.762} & \scalebox{0.78}{0.477} &\scalebox{0.78}{0.629} &{\scalebox{0.78}{0.336}}  &{\scalebox{0.78}{0.605}} &\scalebox{0.78}{0.373} & \scalebox{0.78}{0.797} & \scalebox{0.78}{0.508}&\scalebox{0.78}{0.621} &\scalebox{0.78}{0.383} &\scalebox{0.78}{0.618} &{\scalebox{0.78}{0.328}} &\scalebox{0.78}{0.622} &\scalebox{0.78}{0.337} \\ 
    & \scalebox{0.78}{720} 
    & \boldres{\scalebox{0.78}{0.461}} & \boldres{\scalebox{0.78}{0.297}}
    &\scalebox{0.78}{0.655} &\scalebox{0.78}{0.319}
    &\scalebox{0.78}{0.676} &\scalebox{0.78}{0.327}
    &\scalebox{0.78}{0.518} &\scalebox{0.78}{0.338}& {\scalebox{0.78}{0.534}} & {\scalebox{0.78}{0.342}} & \secondres{\scalebox{0.78}{0.467}} & \secondres{\scalebox{0.78}{0.302}} & \scalebox{0.78}{0.647} & \scalebox{0.78}{0.387} & {\scalebox{0.78}{0.514}} & {\scalebox{0.78}{0.322}} & \scalebox{0.78}{0.589} & \scalebox{0.78}{0.328}  & \scalebox{0.78}{0.719} & \scalebox{0.78}{0.449} &\scalebox{0.78}{0.640} &{\scalebox{0.78}{0.350}} &\scalebox{0.78}{0.645} &\scalebox{0.78}{0.394} & \scalebox{0.78}{0.841} & \scalebox{0.78}{0.523} &{\scalebox{0.78}{0.626}} &\scalebox{0.78}{0.382} &\scalebox{0.78}{0.653} &{\scalebox{0.78}{0.355}} &\scalebox{0.78}{0.660} &\scalebox{0.78}{0.408} \\ 
    \cmidrule(lr){2-34}
    & \scalebox{0.78}{Avg} &\boldres{\scalebox{0.78}{0.422}} & \boldres{\scalebox{0.78}{0.276}}
    &\scalebox{0.78}{0.622} & \scalebox{0.78}{0.309}
    &\scalebox{0.78}{0.670} & \scalebox{0.78}{0.332}
    &\scalebox{0.78}{0.481} & \scalebox{0.78}{0.321}
    &\scalebox{0.78}{0.511} & \scalebox{0.78}{0.334}& \secondres{\scalebox{0.78}{0.428}} & \secondres{\scalebox{0.78}{0.282}} & \scalebox{0.78}{0.626} & \scalebox{0.78}{0.378} & {\scalebox{0.78}{0.481}} & {\scalebox{0.78}{0.304}}& \scalebox{0.78}{0.550} & {\scalebox{0.78}{0.304}} & \scalebox{0.78}{0.760} & \scalebox{0.78}{0.473} &{\scalebox{0.78}{0.620}} &{\scalebox{0.78}{0.336}} &\scalebox{0.78}{0.625} &\scalebox{0.78}{0.383} & \scalebox{0.78}{0.804} & \scalebox{0.78}{0.509} &{\scalebox{0.78}{0.610}} &\scalebox{0.78}{0.376} &\scalebox{0.78}{0.624} &{\scalebox{0.78}{0.340}} &\scalebox{0.78}{0.628} &\scalebox{0.78}{0.379} \\ 
    \midrule
    
    \multirow{4}{*}{\rotatebox{90}{\scalebox{0.95}{Weather}}} 
    &  \scalebox{0.78}{96} 
    & {\scalebox{0.78}{0.165}} & \secondres{{\scalebox{0.78}{0.214}}}
    & \boldres{\scalebox{0.78}{0.160}} &{\scalebox{0.78}{0.228}}
    & {\scalebox{0.78}{0.167}} &{\scalebox{0.78}{0.228}}
    & \secondres{\scalebox{0.78}{0.164}} &\boldres{\scalebox{0.78}{0.212}}
    &\scalebox{0.78}{0.186} & {\scalebox{0.78}{0.227}}& \scalebox{0.78}{0.174} & \secondres{\scalebox{0.78}{0.214}} & \scalebox{0.78}{0.192} & \scalebox{0.78}{0.232} & \scalebox{0.78}{0.177} & {\scalebox{0.78}{0.218}} & {\scalebox{0.78}{0.158}} & \scalebox{0.78}{0.230}  & \scalebox{0.78}{0.202} & \scalebox{0.78}{0.261} &{\scalebox{0.78}{0.172}} &{\scalebox{0.78}{0.220}} & \scalebox{0.78}{0.196} &\scalebox{0.78}{0.255} & \scalebox{0.78}{0.221} & \scalebox{0.78}{0.306} & \scalebox{0.78}{0.217} &\scalebox{0.78}{0.296} & {\scalebox{0.78}{0.173}} &{\scalebox{0.78}{0.223}} & \scalebox{0.78}{0.266} &\scalebox{0.78}{0.336} \\ 
    & \scalebox{0.78}{192} & {\scalebox{0.78}{0.214}} & {{\scalebox{0.78}{0.255}}}
    &\boldres{\scalebox{0.78}{0.207}} &{\scalebox{0.78}{0.272}}
    &{\scalebox{0.78}{0.213}} &{\scalebox{0.78}{0.274}}
    &\secondres{\scalebox{0.78}{0.212}} &\boldres{\scalebox{0.78}{0.253}}& \scalebox{0.78}{0.232} & {\scalebox{0.78}{0.264}}& \scalebox{0.78}{0.221} & \secondres{\scalebox{0.78}{0.254}} & \scalebox{0.78}{0.240} & \scalebox{0.78}{0.271} & \scalebox{0.78}{0.225} & \scalebox{0.78}{0.259} & \scalebox{0.78}{0.206} & \scalebox{0.78}{0.277} & \scalebox{0.78}{0.242} & \scalebox{0.78}{0.298} &{\scalebox{0.78}{0.219}} &{\scalebox{0.78}{0.261}}  & \scalebox{0.78}{0.237} &\scalebox{0.78}{0.296} & \scalebox{0.78}{0.261} & \scalebox{0.78}{0.340} & \scalebox{0.78}{0.276} &\scalebox{0.78}{0.336} & \scalebox{0.78}{0.245} &\scalebox{0.78}{0.285} & \scalebox{0.78}{0.307} &\scalebox{0.78}{0.367} \\ 
    & \scalebox{0.78}{336} & \secondres{{\scalebox{0.78}{0.271}}} & {\scalebox{0.78}{0.297}}
     & {\scalebox{0.78}{0.280}} &{\scalebox{0.78}{0.318}}
      & {\scalebox{0.78}{0.289}} &{\scalebox{0.78}{0.328}}
    & \boldres{\scalebox{0.78}{0.269}} &\boldres{\scalebox{0.78}{0.294}}& {\scalebox{0.78}{0.288}} & \scalebox{0.78}{0.304}& {\scalebox{0.78}{0.278}} & \secondres{\scalebox{0.78}{0.296}}& \scalebox{0.78}{0.292} & \scalebox{0.78}{0.307} & \scalebox{0.78}{0.278} & {\scalebox{0.78}{0.297}} & \scalebox{0.78}{0.272} & \scalebox{0.78}{0.335} & \scalebox{0.78}{0.287} & \scalebox{0.78}{0.335} &{\scalebox{0.78}{0.280}} &{\scalebox{0.78}{0.306}} & \scalebox{0.78}{0.283} &\scalebox{0.78}{0.335} & \scalebox{0.78}{0.309} & \scalebox{0.78}{0.378} & \scalebox{0.78}{0.339} &\scalebox{0.78}{0.380} & \scalebox{0.78}{0.321} &\scalebox{0.78}{0.338} & \scalebox{0.78}{0.359} &\scalebox{0.78}{0.395}\\ 
    & \scalebox{0.78}{720} & \boldres{\scalebox{0.78}{0.346}} & \secondres{\scalebox{0.78}{0.347}}
    &{\scalebox{0.78}{0.350}} &{\scalebox{0.78}{0.374}}
    &{\scalebox{0.78}{0.358}} &{\scalebox{0.78}{0.377}}
    &\secondres{\scalebox{0.78}{0.349}} &\boldres{\scalebox{0.78}{0.346}}& \scalebox{0.78}{0.361} & \scalebox{0.78}{0.351} & \scalebox{0.78}{0.358} & \secondres{\scalebox{0.78}{0.347}}& \scalebox{0.78}{0.364} & \scalebox{0.78}{0.353} & \scalebox{0.78}{0.354} & {\scalebox{0.78}{0.348}} & \scalebox{0.78}{0.398} & \scalebox{0.78}{0.418} & {\scalebox{0.78}{0.351}} & \scalebox{0.78}{0.386} &\scalebox{0.78}{0.365} &{\scalebox{0.78}{0.359}} & \scalebox{0.78}{0.345} &{\scalebox{0.78}{0.381}} & \scalebox{0.78}{0.377} & \scalebox{0.78}{0.427} & \scalebox{0.78}{0.403} &\scalebox{0.78}{0.428} & \scalebox{0.78}{0.414} &\scalebox{0.78}{0.410} & \scalebox{0.78}{0.419} &\scalebox{0.78}{0.428} \\ 
    \cmidrule(lr){2-34}
    & \scalebox{0.78}{Avg} 
    &\boldres{\scalebox{0.78}{0.249}} & \secondres{\scalebox{0.78}{0.278}}
    &\secondres{\scalebox{0.78}{0.250}}
    & {\scalebox{0.78}{0.298}}
    &{\scalebox{0.78}{0.256}}
    & {\scalebox{0.78}{0.302}}
    &\boldres{\scalebox{0.78}{0.249}}
    & \boldres{\scalebox{0.78}{0.276}}
    &\scalebox{0.78}{0.267} & \scalebox{0.78}{0.287}
    & {\scalebox{0.78}{0.258}} & \secondres{\scalebox{0.78}{0.278}} & \scalebox{0.78}{0.272} & \scalebox{0.78}{0.291} & {\scalebox{0.78}{0.259}} & {\scalebox{0.78}{0.281}} & \scalebox{0.78}{0.259} & \scalebox{0.78}{0.315} & \scalebox{0.78}{0.271} & \scalebox{0.78}{0.320} &{\scalebox{0.78}{0.259}} &{\scalebox{0.78}{0.287}} &\scalebox{0.78}{0.265} &\scalebox{0.78}{0.317} & \scalebox{0.78}{0.292} & \scalebox{0.78}{0.363} &\scalebox{0.78}{0.309} &\scalebox{0.78}{0.360} &\scalebox{0.78}{0.288} &\scalebox{0.78}{0.314} &\scalebox{0.78}{0.338} &\scalebox{0.78}{0.382} \\ 
     \midrule
    
    \multirow{5}{*}{\rotatebox{90}{\scalebox{0.95}{Solar-Energy}}} 
    
    &  \scalebox{0.78}{96} 
    &\boldres{\scalebox{0.78}{0.186}} & \boldres{\scalebox{0.78}{0.217}}
    &\scalebox{0.78}{0.195} & \scalebox{0.78}{0.251}
    &\scalebox{0.78}{0.219} & \scalebox{0.78}{0.270}
    &\scalebox{0.78}{0.210} & \scalebox{0.78}{0.249}
    &\scalebox{0.78}{0.244} & \scalebox{0.78}{0.293}
    &\secondres{\scalebox{0.78}{0.203}} &\secondres{\scalebox{0.78}{0.237}} & \scalebox{0.78}{0.322} & \scalebox{0.78}{0.339} & {\scalebox{0.78}{0.234}} & {\scalebox{0.78}{0.286}} &\scalebox{0.78}{0.310} &\scalebox{0.78}{0.331} &\scalebox{0.78}{0.312} &\scalebox{0.78}{0.399} &\scalebox{0.78}{0.250} &\scalebox{0.78}{0.292} &\scalebox{0.78}{0.290} &\scalebox{0.78}{0.378} &\scalebox{0.78}{0.237} &\scalebox{0.78}{0.344} &\scalebox{0.78}{0.242} &\scalebox{0.78}{0.342} &\scalebox{0.78}{0.215} &\scalebox{0.78}{0.249} &\scalebox{0.78}{0.884} &\scalebox{0.78}{0.711}\\ 
    & \scalebox{0.78}{192} 
    &\secondres{\scalebox{0.78}{0.230}} & \boldres{\scalebox{0.78}{0.251}}
    &\boldres{\scalebox{0.78}{0.214}} & \scalebox{0.78}{0.266}
    &\scalebox{0.78}{0.236} & \scalebox{0.78}{0.282}
    &\scalebox{0.78}{0.242} & \scalebox{0.78}{0.274}
    &\scalebox{0.78}{0.265} & \scalebox{0.78}{0.298}
    &{\scalebox{0.78}{0.233}} &\secondres{\scalebox{0.78}{0.261}} & \scalebox{0.78}{0.359} & \scalebox{0.78}{0.356}& {\scalebox{0.78}{0.267}} & {\scalebox{0.78}{0.310}} &\scalebox{0.78}{0.734} &\scalebox{0.78}{0.725} &\scalebox{0.78}{0.339} &\scalebox{0.78}{0.416} &\scalebox{0.78}{0.296} &\scalebox{0.78}{0.318} &\scalebox{0.78}{0.320} &\scalebox{0.78}{0.398} &\scalebox{0.78}{0.280} &\scalebox{0.78}{0.380} &\scalebox{0.78}{0.285} &\scalebox{0.78}{0.380} &\scalebox{0.78}{0.254} &\scalebox{0.78}{0.272} &\scalebox{0.78}{0.834} &\scalebox{0.78}{0.692} \\ 
    & \scalebox{0.78}{336} 
    &{\scalebox{0.78}{0.253}} & \boldres{\scalebox{0.78}{0.270}}
     &\boldres{\scalebox{0.78}{0.245}} & \scalebox{0.78}{0.274}
      &\scalebox{0.78}{0.250} & \scalebox{0.78}{0.288}
    &\scalebox{0.78}{0.260} & \scalebox{0.78}{0.287}
    &\scalebox{0.78}{0.284} & \scalebox{0.78}{0.312}
    &\secondres{\scalebox{0.78}{0.248}} &\secondres{\scalebox{0.78}{0.273}} & \scalebox{0.78}{0.397} & \scalebox{0.78}{0.369}& {\scalebox{0.78}{0.290}}  &{\scalebox{0.78}{0.315}} &\scalebox{0.78}{0.750} &\scalebox{0.78}{0.735} &\scalebox{0.78}{0.368} &\scalebox{0.78}{0.430} &\scalebox{0.78}{0.319} &\scalebox{0.78}{0.330} &\scalebox{0.78}{0.353} &\scalebox{0.78}{0.415} &\scalebox{0.78}{0.304} &\scalebox{0.78}{0.389} &\scalebox{0.78}{0.282} &\scalebox{0.78}{0.376} &\scalebox{0.78}{0.290} &\scalebox{0.78}{0.296} &\scalebox{0.78}{0.941} &\scalebox{0.78}{0.723} \\ 
    & \scalebox{0.78}{720} 
    &\secondres{\scalebox{0.78}{0.247}} & \boldres{\scalebox{0.78}{0.274}}
     &\boldres{\scalebox{0.78}{0.246}} & \boldres{\scalebox{0.78}{0.274}}
      &\scalebox{0.78}{0.251} & \scalebox{0.78}{0.286}
    &\scalebox{0.78}{0.268} & \scalebox{0.78}{0.296}
    &\scalebox{0.78}{0.281} & \scalebox{0.78}{0.312}
    &{\scalebox{0.78}{0.249}} &\secondres{\scalebox{0.78}{0.275}} & \scalebox{0.78}{0.397} & \scalebox{0.78}{0.356} & {\scalebox{0.78}{0.289}} &{\scalebox{0.78}{0.317}} &\scalebox{0.78}{0.769} &\scalebox{0.78}{0.765} &\scalebox{0.78}{0.370} &\scalebox{0.78}{0.425} &\scalebox{0.78}{0.338} &\scalebox{0.78}{0.337} &\scalebox{0.78}{0.356} &\scalebox{0.78}{0.413} &\scalebox{0.78}{0.308} &\scalebox{0.78}{0.388} &\scalebox{0.78}{0.357} &\scalebox{0.78}{0.427} &\scalebox{0.78}{0.285} &\scalebox{0.78}{0.295} &\scalebox{0.78}{0.882} &\scalebox{0.78}{0.717} \\ 
    \cmidrule(lr){2-34}
    & \scalebox{0.78}{Avg} 
    &\secondres{\scalebox{0.78}{0.229}} & \boldres{\scalebox{0.78}{0.253}}
     &\boldres{\scalebox{0.78}{0.225}} & \scalebox{0.78}{0.266}
      &\scalebox{0.78}{0.239} & \scalebox{0.78}{0.282}
    &\scalebox{0.78}{0.245} & \scalebox{0.78}{0.277}
    &\scalebox{0.78}{0.269} & \scalebox{0.78}{0.304}
    &{\scalebox{0.78}{0.233}} &\secondres{\scalebox{0.78}{0.262}} & \scalebox{0.78}{0.369} & \scalebox{0.78}{0.356} &{\scalebox{0.78}{0.270}} &{\scalebox{0.78}{0.307}} &\scalebox{0.78}{0.641} &\scalebox{0.78}{0.639} &\scalebox{0.78}{0.347} &\scalebox{0.78}{0.417} &\scalebox{0.78}{0.301} &\scalebox{0.78}{0.319} &\scalebox{0.78}{0.330} &\scalebox{0.78}{0.401} &\scalebox{0.78}{0.282} &\scalebox{0.78}{0.375} &\scalebox{0.78}{0.291} &\scalebox{0.78}{0.381} &\scalebox{0.78}{0.261} &\scalebox{0.78}{0.381} &\scalebox{0.78}{0.885} &\scalebox{0.78}{0.711} \\ 

     \midrule
     \multicolumn{2}{c|}{\scalebox{0.78}{{$1^{\text{st}}$ Count}}} & \scalebox{0.78}{\boldres{22}} & \scalebox{0.78}{\boldres{25}} &
     \secondres{\scalebox{0.78}{15}} & \scalebox{0.78}{\secondres{7}}& 
     {\scalebox{0.78}{4}} & \scalebox{0.78}{{3}} &
     {\scalebox{0.78}{3}} & \scalebox{0.78}{{6}} & \scalebox{0.78}{2} & \scalebox{0.78}{2} & \scalebox{0.78}{1} & \scalebox{0.78}{3} & \scalebox{0.78}{0} & \scalebox{0.78}{3} & \scalebox{0.78}{1} & \scalebox{0.78}{2} & \scalebox{0.78}{0} & \scalebox{0.78}{0} & \scalebox{0.78}{0} & \scalebox{0.78}{0} & \scalebox{0.78}{0} & \scalebox{0.78}{0} & \scalebox{0.78}{2} & \scalebox{0.78}{0} & \scalebox{0.78}{0} & \scalebox{0.78}{0} & {\scalebox{0.78}{1}}& \scalebox{0.78}{0}& \scalebox{0.78}{0}& \scalebox{0.78}{0}& \scalebox{0.78}{0}& \scalebox{0.78}{0}\\ 
    \bottomrule
  \end{tabular}
    \end{small}
  \end{threeparttable}
}
\end{table*}

\end{document}